\documentclass[11pt]{article}

\usepackage{acl}
\usepackage{float}
\usepackage{times}
\usepackage{latexsym}
\usepackage[T1]{fontenc}
\usepackage[utf8]{inputenc}
\usepackage{microtype}
\usepackage{inconsolata}
\usepackage{graphicx}
\usepackage{amsmath}
\usepackage{amssymb}
\usepackage[table]{xcolor}
\usepackage{placeins}
\usepackage{enumitem}
\usepackage{mathtools}
\mathtoolsset{showonlyrefs=true}
\usepackage{comment}
\usepackage{multirow}

\newcommand{\hh}[1]{{\color{red}[HH: #1]}}
\newcommand{\best}[1]{\textbf{#1}}
\newcommand{\second}[1]{\underline{#1}}

\title{SAC-Copula: Quality-Preserving Watermarking for Diffusion Language Models via Smooth Correlated Gumbel Fields}

\author{
Baixin Li \and Haiyun He \\
The Hong Kong University of Science and Technology (Guangzhou) \\
\texttt{baixinli31@connect.hku.hk, haiyunhe@hkust-gz.edu.cn}
}

\begin{document}
\maketitle

\begin{abstract}
Watermarking diffusion language models (DLMs) requires mechanisms compatible with iterative parallel unmasking rather than autoregressive decoding. Existing sampling-based watermarking methods typically inject position-wise i.i.d.\ perturbations, which can be poorly aligned with DLM decoding dynamics and degrade generation quality. We propose SAC-Copula, a quality-preserving watermarking method for DLMs based on smooth, locally correlated Gumbel perturbation fields constructed via a Gaussian copula. We further develop a SAC-aware detector using covariance-aware filtering and native-sample calibration. Mechanism-level analysis shows that local correlation reduces latent perturbation roughness and better matches iterative refinement dynamics. Experiments on LLaDA show that SAC-Copula achieves a favorable quality--detectability trade-off compared with existing baselines. In particular, further evaluations on Dream-7B and additional datasets show that SAC-Copula substantially improves PPL tail stability over the i.i.d.\ Gumbel baseline, while maintaining strong low-FPR detectability and competitive overall generation quality.
Additional token-edit stress tests further assess watermark robustness under controlled synchronization drift.
Code is available at \url{https://github.com/PunkyKnife/SAC-Copula}.
\end{abstract}

\begin{figure}[t]
    \centering
    \includegraphics[width=\columnwidth]{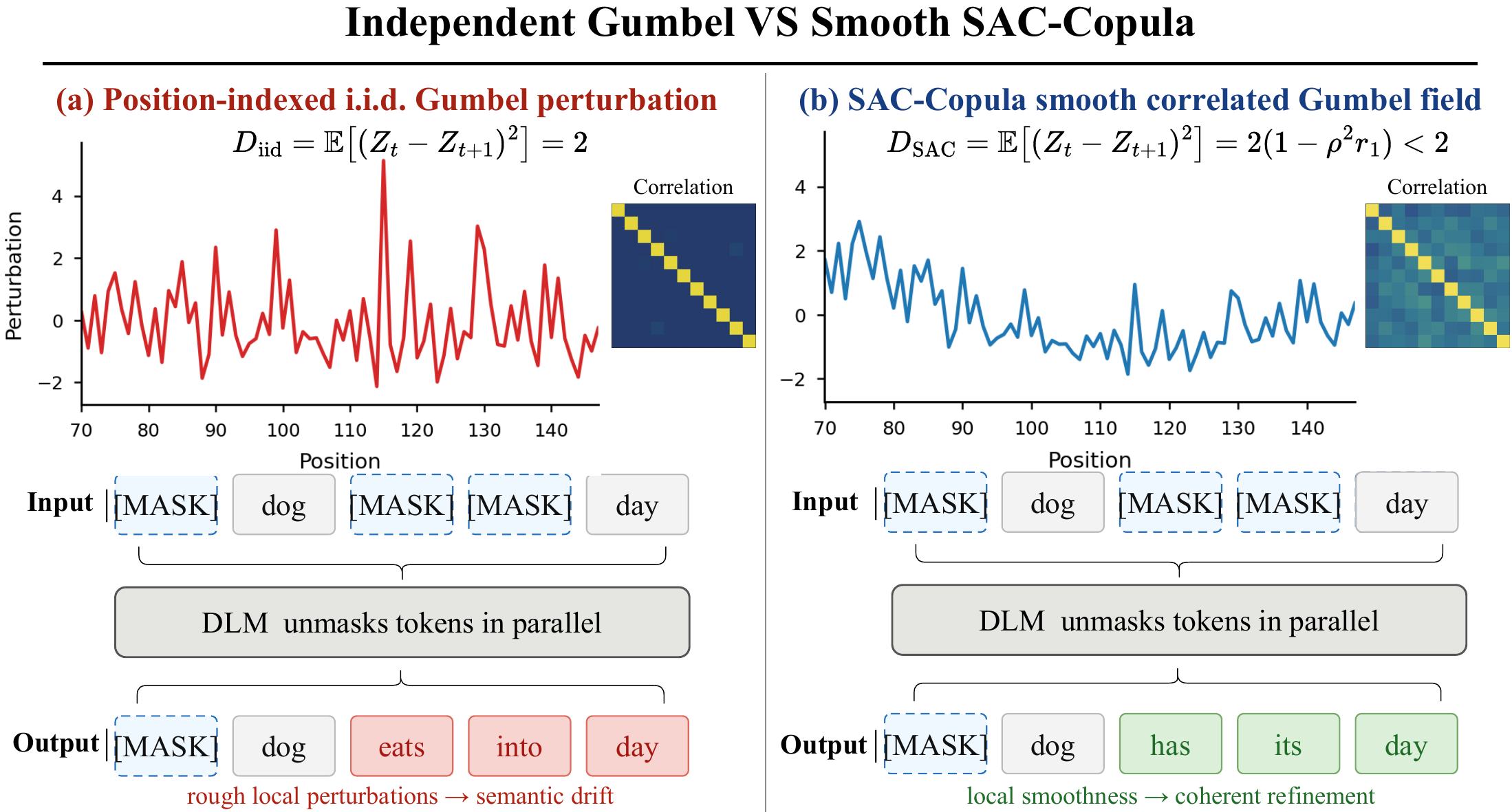}
    \caption{Motivation for smooth watermark fields in DLM decoding.  Position-indexed i.i.d.\ Gumbel perturbations create independent high-frequency noises across neighboring positions, whereas SAC-Copula introduces controlled low-lag dependence while preserving one-coordinate Gumbel marginals.  Token examples are illustrative rather than deterministic generation outcomes.} 
    \label{fig:sac_motivation}
    \vspace{-1em}
\end{figure}

\begin{figure*}[!t]
    \centering
    \includegraphics[width=\linewidth]{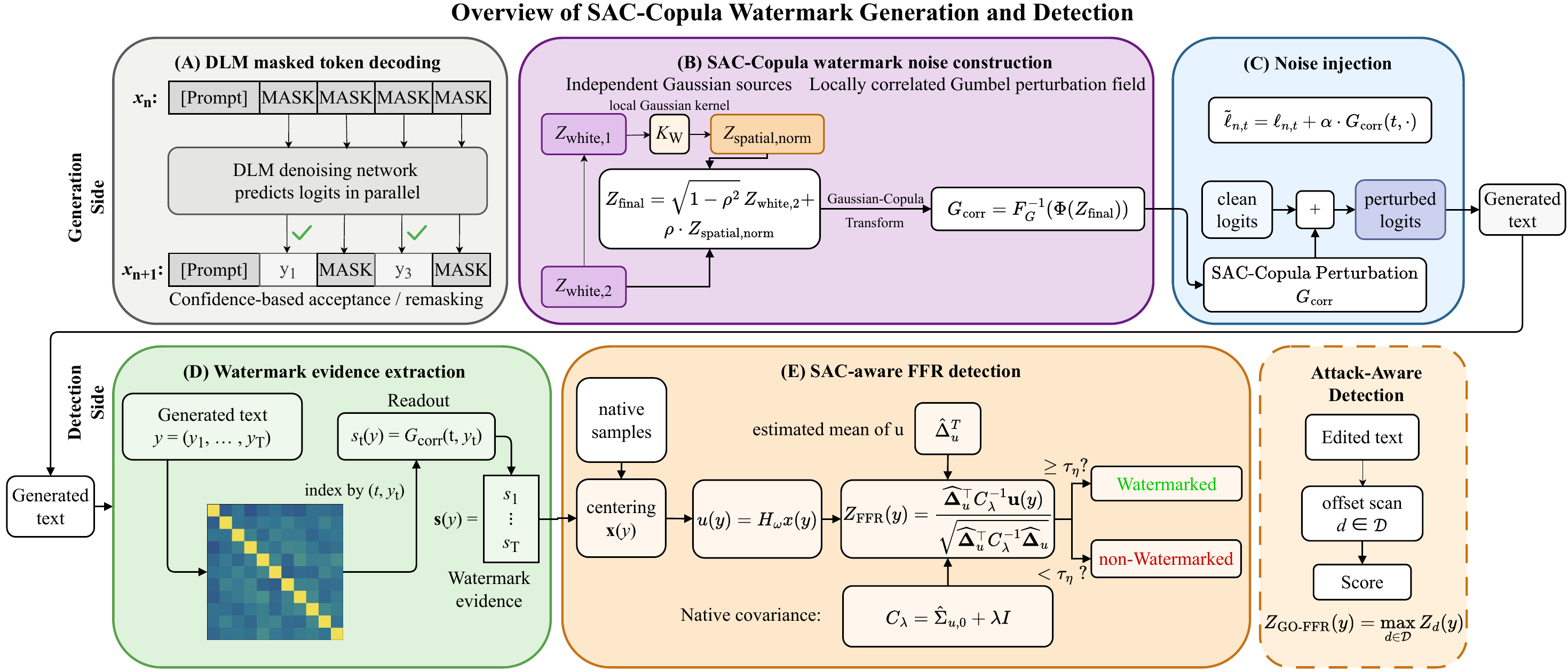}
    \caption{Overview of SAC-Copula.  The generation side constructs a locally correlated Gumbel tape through variance-normalized convolution, dual-stream mixing, and a Gaussian-copula transform, then injects it at currently masked generated positions during DLM refinement.  The detection side reads token-level evidence and applies SAC-aware filtering with thresholds calibrated on native, unwatermarked samples.  GO-FFR is shown only as an auxiliary diagnostic for controlled insertion/deletion edits.}
    \label{fig:method-overview}
\end{figure*}

\section{Introduction}
\label{sec:intro-related}

As Large Language Models (LLMs) become widely deployed, it becomes increasingly critical to establish a safety line between LLMs and humans, where one important brick is to distinguish the model-generated content from the human-created one.  Watermarking offers a practical route: a generator embeds a private statistical signal during content decoding, and a detector later tests whether a text carries that signal via a shared secret key.  

Most existing language-model watermarking methods are developed for autoregressive LLMs, including green-list token biasing, sampling-based or distortion-free schemes, robust watermarking variants, and multi-bit identification mechanisms~\citep{P05_KGW,P14_robust_distortion_free,zhao2023provablerobustwatermarkingaigenerated,yoo2024advancingidentificationmultibitwatermark}. In these methods, watermark signals are typically synchronized with the left context and tokens are generated sequentially from left to right conditioned on the prefix and watermark signal. In contrast, Diffusion Language Models (DLMs), such as LLaDA~\citep{P07_llada}, generate text through iterative masked denoising, where multiple positions are updated in parallel and previously generated tokens can be revised across steps. This non-sequential and dynamically coupled decoding process makes watermark design fundamentally different from the autoregressive setting. In particular, watermarking schemes based on i.i.d.\ injected signals are poorly matched to DLM decoding dynamics, which can disrupt local refinement consistency across denoising steps. This may lead to degraded generation quality, including unstable neighboring predictions, semantic inconsistency, degraded fluency, and refinement oscillation. Figure~\ref{fig:sac_motivation} shows an illustrative example that compares i.i.d.\ and correlated perturbations.


\subsection{Related Works}
\paragraph{Autoregressive and sampling-based watermarking.}
A large body of LLM watermarking work embeds keyed statistical signals during autoregressive decoding and detects them under a false-positive constraint~\citep{P05_KGW,P01_statistical_framework}.  Stronger watermark signals can improve detection power but may degrade fluency, semantic fidelity, or diversity, motivating selective, adaptive, and quality-aware objectives~\citep{P06_learning_to_watermark,P24_watermax,P25_dawa}.  A related line uses Gumbel-max, inverse-transform, or distortion-free sampling ideas to embed keyed randomness under specific sampling and security assumptions~\citep{gumbel2023,P13_publicly_detectable,P19_undetectable}. This paper follows the sampling-based watermarking paradigm, while explicitly adapting the design to DLM decoding dynamics. 

\paragraph{DLM watermarking.}
Some recent works have started to explore designing watermarking methods adapted to DLMs. \citet{P22_gloaguen_dlm_watermark} propose the first watermark tailored for DLMs by applying the green-red list watermarking \citep{P05_KGW} in expectation over partially observed contexts and promoting tokens that strengthen future watermark signals. To address the non-sequential decoding process, some works \citep{P04_dmark, raban2026lrdwm} further consider synchronizing watermark signals using both left and right contexts. 
Instead of directly perturbing token probabilities, \citet{hong2026dgmark} embed watermark by steering the decoding order to improve quality preservation in DLMs. 

More closely related to our work, \citet{P23_ddlm_watermark} apply Gumbel-max watermarking \cite{gumbel2023} to diffusion decoding using sequence-position-indexed randomness. Although this approach is theoretically distribution-preserving, the injected i.i.d.\ Gumbel perturbations are not well aligned with DLM decoding dynamics, as discussed earlier, and can degrade generated text quality in practice. Existing Gumbel-based methods primarily focus on how randomness is indexed across positions. In contrast, we argue that the \emph{joint structure} of the perturbation field is itself a critical design dimension for DLM watermarking. Under iterative refinement decoding, independent high-frequency perturbations across neighboring positions can conflict with local consistency. Motivated by this observation, we shift the design focus from injecting independent randomness to designing structured perturbation fields whose geometry and smoothness better align with diffusion decoding dynamics and detector evidence geometry.

The main insight of our paper is that locally smooth and correlated watermark perturbations may be more compatible with DLM decoding than high-frequency i.i.d.\ perturbations, motivating the hypothesis of improved generation stability. Based on this insight, we propose \textbf{SAC-Copula}, which constructs correlated Gumbel perturbation fields via a Gaussian copula (cf.~Figure \ref{fig:sac_motivation}). Under suitable assumptions, SAC-Copula preserves Gumbel marginals while introducing controlled local dependence across neighboring token positions. This structured perturbation field is designed to better align with DLM refinement dynamics and supports a strong quality--detectability trade-off.

\subsection{Main Contributions}


Our contributions can be summarized as follows:
\begin{itemize}[leftmargin=1em,itemsep=0pt]
    \item We introduce \textbf{SAC-Copula}, a DLM watermarking method with \textbf{S}mooth and \textbf{A}uto-\textbf{C}orrelated Gumbel Copula perturbations through a Gaussian \textbf{Copula}, which preserves one-coordinate Gumbel marginals under mild assumptions. This is a DLM-aware watermarking perspective that treats the joint geometry of the perturbation field as a key design dimension for DLM decoding. 


    \item We provide a mechanism-level analysis showing that positive low-lag dependence reduces latent perturbation roughness, decreasing the local-difference energy from $D_{\mathrm{i.i.d}} = 2$ to $D_{\mathrm{SAC}} = 2(1-\rho^2 r_1)$ and providing a mechanistic rationale for the DLM-refinement hypothesis.

    \item We develop a SAC-aware detector that reads the induced local evidence geometry through covariance-aware filtering and native-sample calibration. For controlled edit diagnostics, we further include a global-offset scan that provides coarse synchronization recovery under token--signal misalignment.

    \item Experiments on LLaDA show that SAC-Copula achieves a strong quality--detectability trade-off, with strong clean detectability and a balanced multi-dimensional quality profile. Relative to the matched i.i.d.\ Gumbel baseline, the clearest generation-side gain is substantially improved upper-tail stability. We further report mild token-edit stress tests as diagnostic analyses of residual watermark evidence.
\end{itemize}

\section{Watermark Injection and Detection}
\label{sec:method}

SAC-Copula replaces position-indexed i.i.d.\ Gumbel perturbations with a smooth, locally correlated Gumbel field for DLM decoding.  The generation side defines this field, while the detector provides a matched readout for the induced evidence geometry.  Figure~\ref{fig:method-overview} summarizes the pipeline.


\paragraph{Baseline i.i.d.\ Gumbel Watermarking.}
Let \(\mathcal V\) be the token vocabulary, \(j\in\mathcal V\) a token, \(t\in[T]\) a generated-token position, and \(n\) a DLM denoising step.  The direct baseline applies position-indexed i.i.d.\ Gumbel perturbations~\citep{P23_ddlm_watermark,gumbel2023},
\(G_{\mathrm{i.i.d.}}(t,j)\sim\mathrm{Gumbel}(0,1)\), to masked-position logits:
\begin{equation*}
\widetilde\ell_{n,t,j}
=
\ell_{n,t,j}
+
\alpha G_{\mathrm{i.i.d.}}(t,j),
\quad t\in[T], j\in\mathcal V.
\end{equation*}
Here \(\ell_{n,t,j}\) is the clean logit and \(\alpha\) is the perturbation scale.  In the controlled mechanism comparison used in this paper, the direct i.i.d.\ Gumbel baseline is the \(\rho=0\) endpoint of the same tape family: setting \(\rho=0\) removes cross-position correlation while retaining standard-Gumbel candidate marginals and the same perturbation scale \(\alpha\). In the main comparison, both settings use \(\alpha=1.0\) and temperature \(1.0\), with the same key/tape interface, prompts, seeds, generation length, diffusion steps, masked-position injection, decoding, and remasking pipeline. We do not introduce or tune a separate \(\delta\)-style logit-bias strength for this matched i.i.d.--SAC comparison; \(\alpha\) is fixed at \(1.0\), while \(W\), \(\sigma_K\), and \(\rho\) parameterize the cross-position dependence structure. 

\subsection{SAC-Copula Perturbation Field}
\label{sec:method-sac}
SAC-Copula constructs a locally correlated Gaussian latent field and maps each coordinate to a standard Gumbel variable through a Gaussian copula.  It starts from two independent key-seeded standard Gaussian position--vocabulary tapes \(Z_{\mathrm{white},1},Z_{\mathrm{white},2}\), with \(Z_{\mathrm{white},r}(t,j)\sim\mathcal N(0,1)\).  Let \(W\) be an odd window, \(h=(W-1)/2\), and \(\sigma_K\) the kernel shape parameter.  For \(k\in\{-h,\ldots,h\}\), define
\begin{equation}
K_W[k]
=
\frac{
\exp\!\left(-k^2/(2\sigma_K^2)\right)
}{
\sum_{m=-h}^{h}\exp\!\left(-m^2/(2\sigma_K^2)\right)
}.
\end{equation}
For full-window positions, the normalized spatial branch is
\begin{equation}
\widetilde Z_{\mathrm{spatial}}(t,j)
=
\frac{
\sum_{k=-h}^{h}K_W[k]Z_{\mathrm{white},1}(t-k,j)
}{
\sqrt{\sum_{k=-h}^{h}K_W[k]^2}
}.
\end{equation}
The final latent field mixes the independent white noise and the normalized spatial stream:
\begin{equation}
\begin{aligned}
Z_{\mathrm{final}}(t,j)
&=
\sqrt{1-\rho^2}\,
Z_{\mathrm{white},2}(t,j)
\\
&\quad+
\rho\,\widetilde Z_{\mathrm{spatial}}(t,j),
\end{aligned}
\end{equation}
where \(\rho\in[0,1]\) controls the local dependence strength.  Under the full-window or boundary-normalized convention, this normalization keeps each valid coordinate of \(Z_{\mathrm{final}}\) standard normal.  Appendices~\ref{app:watermark-boundary} and~\ref{app:watermark-marginal} state the boundary convention and prove the resulting one-coordinate marginal exactness. 

The correlated Gumbel noise tape is obtained by the Gaussian-copula transform
\begin{equation}
G_{\mathrm{corr}}(t,j)
\!=\!
F_G^{-1}\!(\Phi(Z_{\mathrm{final}}(t,j)))\!\sim\! \operatorname{Gumbel}(0,1),
\label{eq:copula-transform}
\end{equation}
where \(\Phi\) is the standard Gaussian cumulative distribution function (CDF) and \(F_G\) is the standard Gumbel CDF. Appendix~\ref{app:watermark-marginal} shows that this coordinate-wise transform preserves one-coordinate Gumbel marginals under assumptions stated in Appendix~\ref{app:watermark-boundary}, while introducing the joint noise tape structure.  
Since the perturbations are also independent across vocabulary items at each fixed position, it further implies that the single-position conditional sampling marginal is preserved:
\[
P_{Y_t\mid\mathrm{context}}^{\mathrm{SAC}}
=
P_{Y_t\mid\mathrm{context}}^{\mathrm{i.i.d.}},
\]
where $Y_t$ denotes the token at the $t$-th position and the context denotes all information used to compute the current logits. This is a fixed-context, single-position statement, not invariance of the full generated-text joint distribution.
Appendix~\ref{app:watermark-autocorr} derives the finite-window latent autocorrelation used by the detector design.

\subsection{Injection into LLaDA Decoding}
\label{sec:method-llada}

SAC-Copula noise is injected only into currently masked generated positions.  At a masked generated position \(t\), SAC-Copula perturbs logits before applying the decoding rule:
\begin{equation}
\widetilde\ell_{n,t,j}
=
\ell_{n,t,j}
+
\alpha G_{\mathrm{corr}}(t,j),
\quad t\in[T], j\in\mathcal V,
\end{equation}
Here \(\ell_{n,t,j}\) is the contemporaneous clean logit from the current partially masked denoising state. The keyed tape \(G_{\mathrm{corr}}\) is constructed independently of model logits; the logits determine how the fixed tape perturbation affects token selection, not how the tape itself is constructed.  The perturbation scale \(\alpha\) is an experimental configuration parameter.  In the main configuration it matches the generation temperature, but this equality is not a theoretical requirement.  Prompt tokens are not perturbed, and remasking confidence is computed from the unperturbed logits.  The same token-position perturbation \(G_{\mathrm{corr}}(t,\cdot)\) is reused across denoising steps, indexed by generated token position and token but not by denoising step.

\subsection{SAC-Aware Detector}
\label{sec:method-detection}

For a clean fixed-alignment sequence \(y=(y_1,\ldots,y_T)\), the detector reconstructs the same private keyed \(G_{\mathrm{corr}}\) tape and indexes it with the observed token at each generated position, yielding token-level watermark evidence \(\mathbf{s}(y)=(s_1(y),\ldots,s_T(y))^\top\), where
\begin{equation}
s_t(y)=G_{\mathrm{corr}}(t,y_t).
\end{equation}
The tape reconstruction depends on the private key rather than on a second model evaluation; detection does not recompute fully revealed logits or require an additional language-model forward pass.
The legacy detector treats these values as exchangeable i.i.d.\ Gumbel evidence and applies an equal-weight z-score; Appendix~\ref{app:detector-covcal} gives the full old-score form.  SAC-Copula intentionally induces low-lag evidence structure, so relying only on an i.i.d.\ equal-weight readout can discard useful local geometry. 
We therefore use the Full Filtered Ridge (FFR) detector, defined as a ridge-regularized matched linear statistic after SAC-aware local filtering and covariance calibration on native samples.

FFR first centers raw evidence using the Gumbel mean \(\mu_G\) and a residual mean $\widehat{\boldsymbol{\mu}}_{x,0}$ estimated from held-out native (aka.~unwatermarked) samples generated under the same decoding protocol:
\begin{equation}
\mathbf{x}(y)
=
\mathbf{s}(y)-\mu_G\mathbf 1-\widehat{\boldsymbol{\mu}}_{x,0}.
\end{equation}
It then applies a SAC-aware local feature map
\begin{equation}
\begin{gathered}
\mathbf u(y)
=
H_\omega \mathbf x(y),
\\
\omega_\tau
\propto
\exp\!\left(-\frac{\tau^2}{4\sigma_K^2}\right)
\mathbf 1\{|\tau|\le B\},
\end{gathered}
\label{eq:local-filter-map}
\end{equation}
where \(H_\omega\) is a filter motivated by the autocorrelation envelope of the SAC kernel (cf.~Appendix \ref{app:detector-filtering}) and \(B=\min(W-1,T-1)\).  The lag template mirrors the Gaussian-envelope smoothing used by SAC-Copula, so \(H_\omega\) aggregates evidence over the local low-lag structure introduced by the correlated perturbation field.

The filtered native-sample covariance and ridge covariance are
\begin{equation}
\widehat\Sigma_{u,0}
=
H_\omega\widehat\Sigma_{x,0}H_\omega^\top,
\qquad
C_\lambda=\widehat\Sigma_{u,0}+\lambda I .
\end{equation}
Here \(\widehat\Sigma_{x,0}\) is estimated from held-out native samples under the same detector configuration. Let \(\widehat{\boldsymbol{\Delta}}_u\) be the filtered mean direction estimated from development or cross-fit data, not evaluation samples.  The FFR score is
\begin{equation}
    Z_{\mathrm{FFR}}(y)=
    \frac{
    \widehat{\boldsymbol{\Delta}}_u^\top
    C_\lambda^{-1}
    \mathbf{u}(y)
    }{
    \sqrt{\widehat{\boldsymbol{\Delta}}_u^\top C_\lambda^{-1} \widehat{\boldsymbol{\Delta}}_u}.
    }
\end{equation}
The detector flags a sequence as watermarked when
\begin{equation}
Z_{\mathrm{FFR}}(y)\ge \tau_\eta,
\end{equation}
where the threshold \(\tau_\eta\) can be calibrated from held-out native samples.  In our experiments, we report ROC-based AUC and TPR at target FPR levels using the corresponding Native-score distribution.  FFR provides a matched linear readout for the SAC-induced evidence geometry under the stated calibration model.  Appendix~\ref{app:detector-covcal}--\ref{app:detector-old-relation}
provide the full FFR rationale, matched-direction derivation, and
relation to the old detector. 

Furthermore, as one step towards robust detection, we introduce another detector, GO-FFR, extended from FFR by considering controlled insertion/deletion attacks:
\begin{equation}
Z_{\mathrm{GO\text{-}FFR}}(Y)
=
\max_{d\in\mathcal D} Z_d(Y),
\end{equation}
where $\mathcal D$ is a scan window and $Z_d(Y)$ is a test statistic defined in Appendix~\ref{app:detector-go}.
It scans a small family of global offsets to partially recover mismatched token–signal alignment, reuses the FFR scoring direction, and calibrates its threshold on native samples under the same offset-scan rule.  
Appendices~\ref{app:detector-go}--\ref{app:detector-scope} give the full analyses.

\section{Why Correlated Perturbations Help DLMs}
\label{sec:why-correlated}

This section provides a mechanism-level rationale for why smooth, locally correlated perturbations may be better aligned with DLM refinement dynamics than high-frequency i.i.d.\ perturbations. We analyze the geometry of the latent perturbation field and the induced detector evidence, and derive empirical predictions on generation quality and detectability that are evaluated in Section~\ref{sec:experiments}.

\subsection{DLM Decoding Couples Neighboring Decisions}
\label{sec:why-dlm-coupling}
Autoregressive watermarking perturbs a left-to-right sampler whose prefix is fixed once a token is emitted.  DLM decoding is different.  LLaDA-style generation repeatedly predicts multiple masked positions in parallel, accepts or remasks tokens according to confidence, and revises a partially generated sequence over denoising steps \citep{P07_llada,P08_mask_predict}.  More recent masked-diffusion and non-Markovian discrete-diffusion work likewise emphasizes iterative denoising and refinement-sensitive generation dynamics \citep{P12_non_markovian_diffusion,P35_masked_diffusion_lm}.  As a result, neighboring token positions are not isolated one-step decisions: local choices can be compared, retained, or revised together during refinement.

This difference makes the geometry of the perturbation field relevant.  Beyond its marginal strength at each position, a watermark also determines how perturbations vary across adjacent positions.  An i.i.d.\ Gumbel field can vary sharply across neighboring masked positions that are refined together, creating a plausible mismatch with the local consistency preference of iterative masked refinement.  A smoother locally correlated field makes nearby perturbation preferences more coherent, which motivates the quality-stability hypothesis tested in Section~\ref{sec:experiments}.
Figure~\ref{fig:sac_motivation} illustrates this contrast, while the following subsections formalize it using a latent local-difference proxy.

\subsection{Independent Latent Fields Create Local Roughness}
\label{sec:why-i.i.d-rough}

We first define a latent local-difference proxy before the coordinate-wise Gumbel transform.  Fix a vocabulary item \(j\), and consider the Gaussian latent field \(Z(t,j)\) that is later mapped monotonically into Gumbel noise.  We measure lag-one local variation by
\begin{equation}
D
=
\mathbb E\!\left[
\left(
Z(t+1,j)-Z(t,j)
\right)^2
\right],
\end{equation}
A larger \(D\) indicates sharper adjacent changes in the latent perturbation field.

For an i.i.d.\ latent Gaussian field, $\operatorname{Var}(Z(t,j))=1$ and $\operatorname{Cov}(Z(t,j),Z(t+1,j))=0$.
Therefore, the difference proxy for i.i.d.\ latent Gaussian field is
\begin{equation}
D_{\mathrm{i.i.d}}
=
2.
\end{equation}

\subsection{SAC-Copula Reduces Local Perturbation Variation}
\label{sec:why-sac-smooth}

SAC-Copula changes the joint structure of the perturbation field rather than the one-coordinate marginal.  Let \(r_\tau(W,\sigma_K)\) denote the finite-window autocorrelation of the normalized spatial branch defined in Appendix~\ref{app:watermark-autocorr}.  This quantity is large at small lags when the smoothing kernel strongly overlaps with its shifted copy, and decays as the lag exceeds the kernel support.  For the final latent field, the nonzero-lag covariance is
\[
\begin{split}
&\operatorname{Cov}\!\left(
Z_{\mathrm{final}}(t,j),
Z_{\mathrm{final}}(t+\tau,j)
\right) =
\rho^2 r_\tau(W,\sigma_K).
\end{split}
\]
Let \(r_1 = r_{\tau=1}(W,\sigma_K)\), and write
\(\Delta_t=Z_{\mathrm{final}}(t+1,j)-Z_{\mathrm{final}}(t,j)\),
the lag-one local difference energy becomes
\begin{equation}
\begin{aligned}
D_{\mathrm{SAC}}
&=
\mathbb E[\Delta_t^2]=
2(1-\rho^2 r_1).
\end{aligned}
\end{equation}
When \(r_1>0\) and \(\rho>0\), SAC-Copula has lower latent local variation than the i.i.d.\ field:
\begin{equation*}
D_{\mathrm{SAC}}
<
D_{\mathrm{i.i.d}}.
\end{equation*}
This lower local variation gives a plausible quality mechanism for DLM decoding.  LLaDA-style models refine neighboring masked positions jointly, so when clean logits are locally uncertain, abrupt independent perturbation changes can push nearby positions toward inconsistent random preferences.  SAC-Copula does not reduce the single-coordinate probability of large Gumbel values: after the monotone Gaussian-copula-to-Gumbel transform, each coordinate still has the standard Gumbel marginal under the assumptions in Appendix~\ref{app:watermark-marginal}.  Instead, it reduces abrupt local changes across positions by inducing positive low-lag dependence.  This motivates the empirical hypothesis that smoother perturbation fields may improve quality stability under iterative masked refinement.

Because the inverse-Gumbel transform is nonlinear, this latent local-difference proxy characterizes only the dependence structure before the copula transform and should not be interpreted as invariance of the full generated-text distribution.

\begin{figure}[tbp]
    \centering
    \includegraphics[width=\columnwidth]{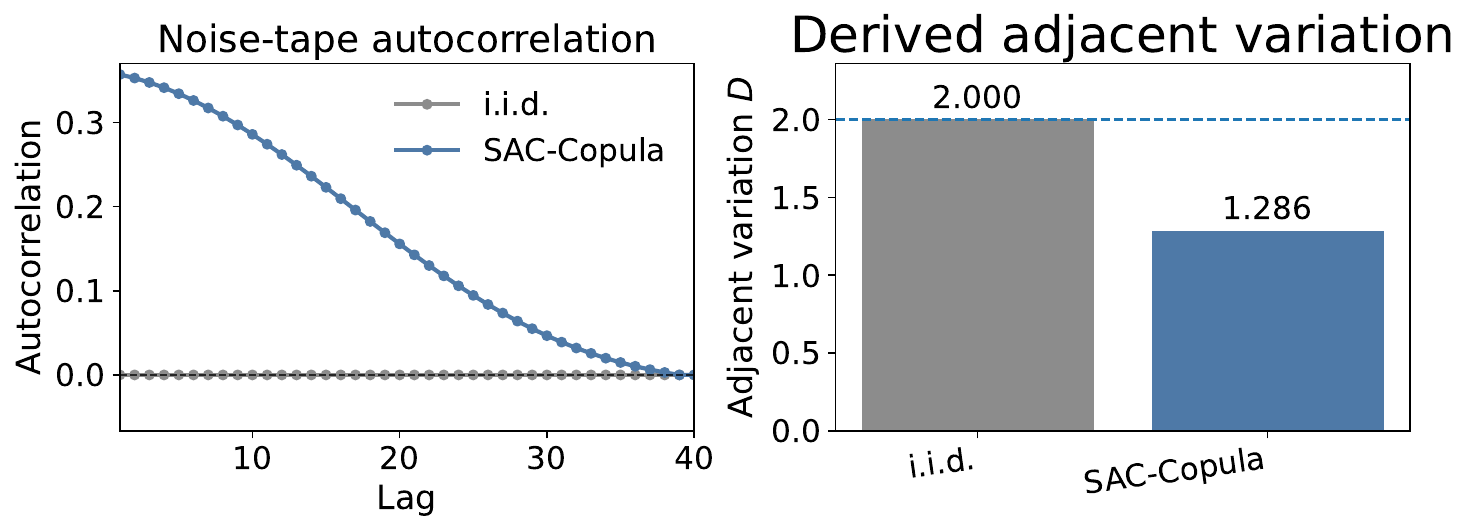}
    \caption{Perturbation-field smoothness diagnostics.  SAC-Copula induces positive low-lag autocorrelation and lowers the latent adjacent-variation proxy from \(D_{\mathrm{iid}}=2\) to
\(D_{\mathrm{SAC}}\approx1.286\).  These are latent-field diagnostics, not text-quality measurements.} 
    \label{fig:section3-smoothness-diagnostics}
\end{figure}

Figure~\ref{fig:section3-smoothness-diagnostics} visualizes these perturbation-field diagnostics.  The autocorrelation view shows that SAC-Copula introduces a positive low-lag dependence, while the adjacent-variation view shows the corresponding reduction in the latent roughness proxy. 


Together, these results suggest that locally correlated perturbations better support DLM decoding quality while preserving watermark detectability. Section~\ref{sec:experiments} evaluates these predictions empirically.

\section{Experiments}
\label{sec:experiments}

We evaluate four questions: clean quality--detectability and matched-i.i.d.\ tail stability; correlation strength; detector, calibration, and targeted-transfer controls; and residual evidence under controlled token edits. 

\subsection{Experimental Setup}
\label{sec:exp-setup}

We evaluate generation from LLaDA-8B-Instruct~\citep{P07_llada} on ELI5 prompts~\citep{fan-etal-2019-eli5}. LLaDA--ELI5 is our primary benchmark; the targeted Dream-7B and C4-en evaluations reported below test transfer across a second DLM backbone and a second task/source setting without redefining the primary protocol. 
The comparison includes native decoding and the direct i.i.d.\ Gumbel baseline.
We evaluate the LLaDA-adapted KGW-style~\citep{P05_KGW} and
Unigram~\citep{zhao2023provablerobustwatermarkingaigenerated} baselines,
together with a PatternMark baseline~\citep{chen2024watermarkorderagnosticlanguagemodels},
over \(\delta\in\{1,2,3\}\).
For KGW and Unigram, we use green-list fraction \(\gamma=0.25\);
each method uses its mechanism-specific detector.

Unless otherwise stated, SAC-Copula uses \(W=39,\sigma_K=15.0,\rho=0.6\), controlling correlation range, kernel shape, and correlated-branch strength. The original mixed-parameter study informed this submitted setting (Appendix~\ref{app:sac-parameter-sensitivity}); the fixed-\((W,\sigma_K)\) sweep below directly tests \(\rho\) and supports \(\rho=0.6\) as a balanced interior operating point.
Full generation, watermark, detector, calibration, seed, and evaluator details are reported in Appendix~\ref{app:full-experimental-configuration}.

Quality metrics (details in Table \ref{tab:app-metric-definitions}) include median PPL, PPL\(>100\), SBERT, MAUVE~\citep{P09_mauve}, Token TV, Rep3, and Distinct3; the matched i.i.d.--SAC comparison additionally reports PPL tail collapse.
AUC and TPR at nominal 1\%/5\% FPR summarize score discrimination~\citep{P01_statistical_framework,P25_dawa}; realized FPR under frozen thresholds tests operational transfer. 

\begin{figure}[tbp]
\centering
\includegraphics[width=0.90\columnwidth]{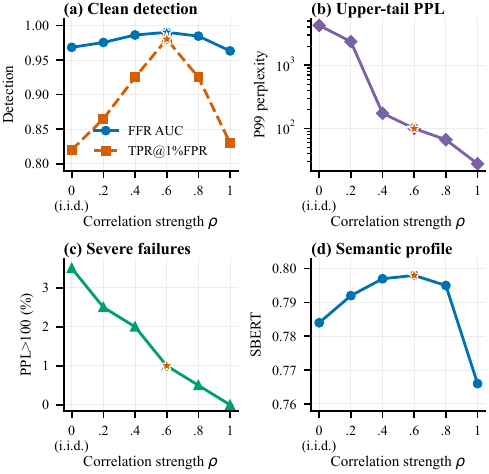}
\caption{Correlation-strength $\rho$ sweep under (\(W=39,\sigma_K=15,\alpha=1\)); \(\rho=0\) is matched i.i.d.\ Gumbel. Detection peaks at moderate correlation and falls at \(\rho=1\); the PPL tail keeps improving, but broader endpoint quality degrades. \(\rho=0.6\) is a balanced interior point.}
\label{fig:rho-sweep}
\end{figure}

\paragraph{Ablation on Correlation Strength.}
Figure~\ref{fig:rho-sweep} varies only \(\rho\) under fixed \(W=39,\sigma_K=15,\alpha=1\), with \(\rho=0\) representing the i.i.d.\ Gumbel. FFR AUC/TPR@1\%FPR rises from \(0.9683/0.820\) at \(\rho=0\) to \(0.9900/0.980\) at \(0.6\), then falls to \(0.9631/0.830\) at \(1\). P99 decreases from \(4256.63\) to \(99.65\) and \(27.55\), but \(\rho=1\) worsens semantic, repetition, and diversity metrics relative to \(0.6\). Thus \(\rho=0.6\) is a balanced interior point; full tables are in Appendix~\ref{app:tail-rho-controls}.

\subsection{Text Quality and Stability}
\label{sec:exp-quality}

\begin{figure}[!t]
\centering
\includegraphics[width=0.79\columnwidth]{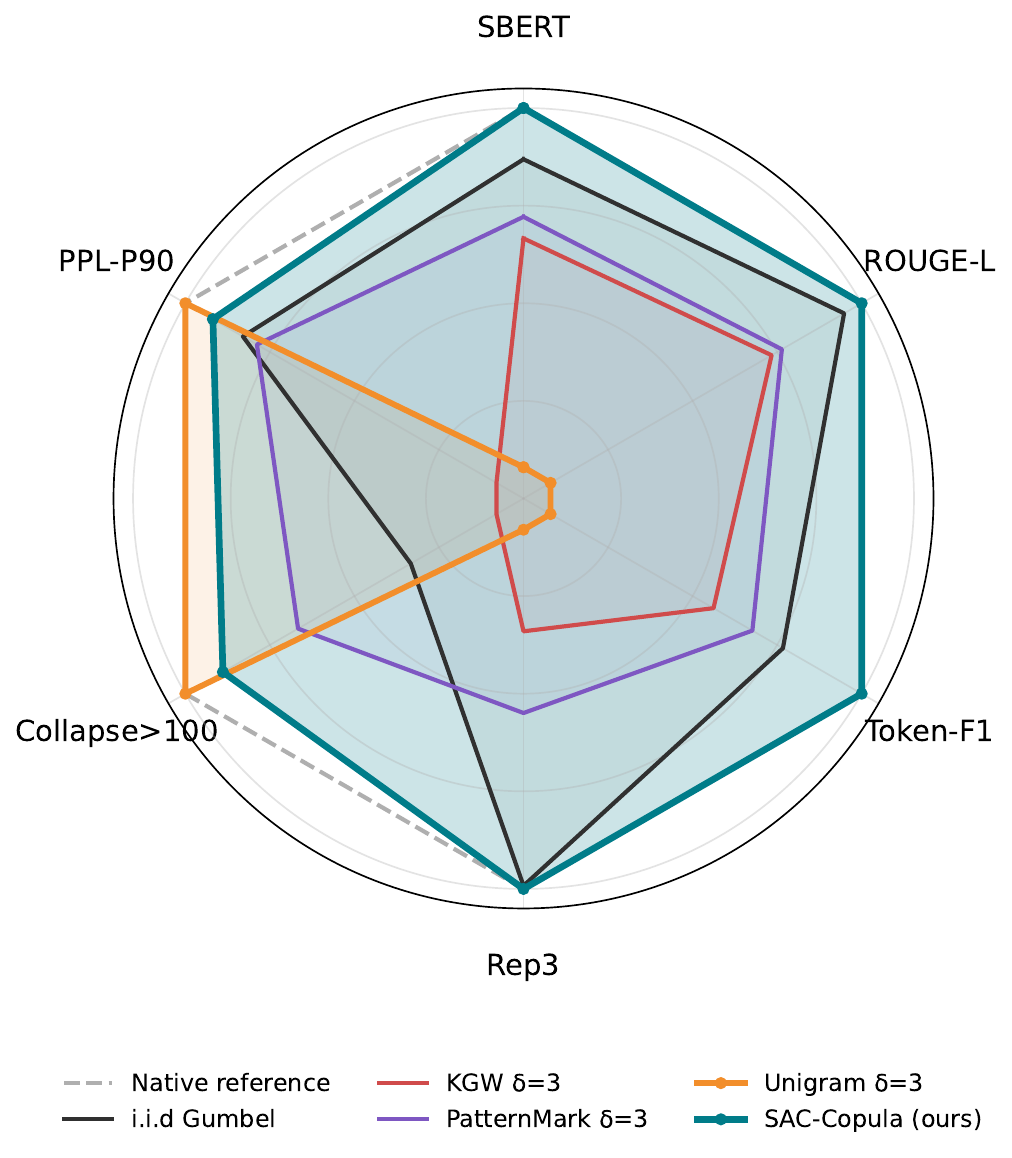}
\caption{Generation-quality profile for SAC-Copula, i.i.d.\ Gumbel, and representative external baselines on direction-normalized axes; larger values indicate better or more native-like behavior. 
}
\label{fig:quality-radar}
\end{figure}

Text quality is the primary empirical target. Relative to matched i.i.d.\ Gumbel, the central PPL values are close (median \(8.446\) vs.\ \(8.260\)), whereas the upper tail differs substantially: P95 decreases from \(48.420\) to \(30.810\), P99 from \(4256.634\) to \(99.653\), and the fraction above PPL 100 from \(3.5\%\) (7/200) to \(1.0\%\) (2/200). Thus, the clearest PPL gain is reduced severe upper-tail failures rather than a uniform per-prompt improvement. Table~\ref{tab:quality-detection} and Figure~\ref{fig:quality-radar} provide the broader context: SAC retains strong low-FPR detectability together with a competitive semantic, distributional, repetition, diversity, and token-drift profile. Full tail diagnostics and complete sweep results are in Appendix~\ref{app:tail-rho-controls}.

\begin{table*}[tbp]
\centering
\scriptsize
\setlength{\tabcolsep}{3pt}
\resizebox{\textwidth}{!}{%
\begin{tabular}{lccccrrrrrrrr}
\hline
Method & Strength 
& AUC $\uparrow$ 
& TPR@1\%FPR $\uparrow$ 
& TPR@5\%FPR $\uparrow$ 
& Median PPL $\downarrow$ 
& PPL\(>100\) $\downarrow$ 
& SBERT $\uparrow$ 
& MAUVE $\uparrow$ 
& Token TV $\downarrow$ 
& Rep3 $\downarrow$ 
& Distinct3 $\uparrow$ 
& Quality $\uparrow$ \\
\hline
Native & -- & -- & -- & -- & 6.381 & 0.005 & 1.000 & 1.000 & 0.000 & 0.406 & 0.594 & 0.960 \\
i.i.d.\ Gumbel & -- & 0.966 & 0.815 & 0.855 & 8.446 & 0.035 & 0.784 & 0.991 & 0.163 & 0.415 & 0.585 & 0.661 \\
\hline
KGW & \(\delta=1\) & 0.670 & 0.060 & 0.120 & 6.697 & 0.030 & 0.794 & \best{0.999} & \best{0.146} & \second{0.414} & \second{0.586} & 0.694 \\
KGW & \(\delta=2\) & 0.820 & 0.200 & 0.315 & 8.020 & 0.045 & 0.776 & \second{0.998} & 0.173 & 0.450 & 0.550 & 0.553 \\
KGW & \(\delta=3\) & 0.936 & 0.515 & 0.660 & 10.543 & 0.050 & 0.762 & 0.982 & 0.228 & 0.510 & 0.489 & 0.379 \\
Unigram & \(\delta=1\) & 0.896 & 0.385 & 0.555 & \best{6.145} & \best{0.000} & \best{0.803} & 0.981 & 0.233 & \best{0.395} & \best{0.605} & \best{0.798} \\
Unigram & \(\delta=2\) & 0.988 & 0.850 & 0.930 & \second{6.535} & \best{0.000} & 0.755 & 0.841 & 0.360 & 0.455 & 0.545 & 0.545 \\
Unigram & \(\delta=3\) & \best{0.997} & \best{0.985} & \best{0.995} & 6.585 & \second{0.005} & 0.690 & 0.601 & 0.439 & 0.560 & 0.440 & 0.150 \\
PatternMark & \(\delta=1\) & 0.818 & 0.510 & 0.625 & 6.766 & 0.010 & 0.781 & 0.983 & \second{0.153} & 0.426 & 0.574 & 0.721 \\
PatternMark & \(\delta=2\) & 0.926 & 0.760 & 0.800 & 7.263 & 0.015 & 0.779 & 0.996 & 0.161 & 0.422 & 0.578 & 0.714 \\
PatternMark & \(\delta=3\) & 0.938 & 0.760 & 0.805 & 8.938 & 0.020 & 0.768 & 0.982 & 0.198 & 0.480 & 0.520 & 0.555 \\
\rowcolor{gray!10}
SAC-Copula (Ours) & -- & \second{0.990} & \second{0.980} & \second{0.985} & 8.260 & 0.010 & \second{0.798} & \best{0.999} & 0.163 & \second{0.414} & \second{0.586} & \second{0.758} \\
\hline
\end{tabular}
}
\caption{Clean detection and text-quality trade-off. AUC and low-FPR TPR report detection; Median PPL and PPL\(>100\) report central fluency and severe tail failures; the remaining raw metrics report broader quality. The normalized Quality score is descriptive only. \best{Best} and \second{second-best} watermarked results are marked in each column.}
\label{tab:quality-detection}
\end{table*}


\subsection{Clean Detection and Calibration}
\label{sec:exp-clean-detection}

Table~\ref{tab:quality-detection} establishes clean fixed-alignment detectability; Appendix~\ref{app:clean_quality_details}, Figure~\ref{fig:app-clean-detection-sweep}, retains the full external sweep.


\paragraph{Detector-readout control.} Under the legacy readout, i.i.d.\ versus SAC AUC/TPR@5\%FPR is \(0.966/0.855\) versus \(0.987/0.950\); FFR further raises SAC TPR@1\%FPR/TPR@5\%FPR to \(0.980/0.985\). Thus the SAC signal difference is visible under the common legacy readout, while FFR provides an additional covariance-aware readout gain. Appendix~\ref{app:detector-comparison}, Table~\ref{tab:sac-detector-comparison}, gives the full control.

\paragraph{Calibration sensitivity.} Table~\ref{tab:calibration-size-main} condenses the model-native \(H_0\) size controls from Appendix~\ref{app:calibration-controls}: score discrimination remains strong and improves with \(N\), while realized FPR at frozen nominal thresholds is operationally sensitive to calibration size. The four-source control in Table~\ref{tab:calibration-source-main} shows only modest changes in score discrimination under this fixed protocol, quantifying calibration-source sensitivity. Threshold portability is nevertheless source-sensitive: Wikipedia-\(H_0\) thresholds yield realized FPRs of \(6.5\%\) and \(21.0\%\) on ELI5 Native at nominal \(1\%\) and \(5\%\), respectively, despite strong score discrimination. Freezing the original LLaDA--ELI5 thresholds without C4 refitting yields \(0.95\%\)/\(5.50\%\) realized FPR on \(2{,}000\) held-out C4 \texttt{realnewslike} negatives at nominal \(1\%\)/\(5\%\); this is detector-only threshold transfer, not C4 generation. Full size, source, prompt, and threshold controls are in Tables~\ref{tab:app-cal-size-roc}--\ref{tab:app-frozen-c4-fpr}.

\begin{table}[tbp]
\centering
\scriptsize
\setlength{\tabcolsep}{3pt}
\renewcommand{\arraystretch}{1.08}
\begin{tabular}{@{}crrrr@{}}
\hline
\(N\) & AUC & TPR@1\%FPR & \shortstack{FPR@nom.~1\%} & \shortstack{FPR@nom.~5\%} \\
\hline
50  & \(.9758\!\pm\!.0111\) & \(.8025\!\pm\!.0715\) & 28.88\% & 34.33\% \\
100 & \(.9867\!\pm\!.0034\) & \(.9078\!\pm\!.0473\) & 13.03\% & 19.55\% \\
200 & \(.9888\!\pm\!.0009\) & \(.9487\!\pm\!.0277\) & 5.13\%  & 11.15\% \\
500 & .9900 & .9800 & 0.50\% & 8.50\% \\
\hline
\end{tabular}
\caption{Calibration-size $N$ controls. AUC and TPR@1\% are mean\(\pm\)SD for repeated \(N<500\) subsamples; the last columns report realized FPR at nominal 1\%/5\% thresholds.}
\label{tab:calibration-size-main}
\end{table}

\begin{table}[tbp]
\centering
\scriptsize
\renewcommand{\arraystretch}{1.08}
\begin{tabular*}{\columnwidth}{
    @{\extracolsep{\fill}}
    l
    @{\hspace{1.5em}}
    r
    @{\hspace{1.5em}}
    r
    @{\hspace{1.5em}}
    r
    @{\extracolsep{\fill}}
}
\hline
\(H_0\) source & AUC & TPR@1\%FPR & TPR@5\%FPR \\
\hline
Native & 0.9900 & 0.980 & 0.985 \\
ELI5 Human & 0.9878 & 0.955 & 0.980 \\
C4 & 0.9871 & 0.955 & 0.975 \\
Wikipedia & 0.9861 & 0.950 & 0.970 \\
\hline
\end{tabular*}
\caption{Four-source \(H_0\) calibration control at \(N=500\) with fixed development and evaluation identities.}
\label{tab:calibration-source-main}
\vspace{-2em}
\end{table}


\paragraph{Targeted model/task transfer.} Table~\ref{tab:targeted-transfer} evaluates SAC-Copula on two additional model–dataset settings. On Dream-7B/ELI5, SAC-Copula maintains near-perfect detection while substantially reducing P99 and the Composite-collapse rate. On LLaDA/C4-en, it improves detection while reducing P99 and the PPL-tail failure rate, with nearly unchanged median PPL (13.71 vs.\ 13.73). These targeted checks extend the evidence to a second backbone and task/source setting without establishing universal transfer. Appendix~\ref{app:targeted-transfer} provides full Dream-7B/ELI5 (Table~\ref{tab:app-dream-transfer}) and LLaDA/C4-en results, including a separate common-detector control (Tables~\ref{tab:app-c4-method-matched}–\ref{tab:app-c4-common-detector}).

\begin{table}[!htbp]
\centering
\scriptsize
\setlength{\tabcolsep}{1.8pt}
\renewcommand{\arraystretch}{1.1}
\begin{tabular}{@{}l|lrrrr@{}}
\hline
\shortstack{Model/\\Dataset} & Method & AUC $\uparrow$ & \shortstack{TPR@1\%FPR $\uparrow$} & P99 $\downarrow$ & \shortstack{Tail failure $\downarrow$} \\
\hline
\multirow{2}{*}{\shortstack[l]{Dream-7B/\\ELI5}}
    & i.i.d.\ Gumbel & 1.0000 & 1.000 & 5988.25 & 24.5\% \\
    & \cellcolor{gray!10} SAC-Copula
    & \cellcolor{gray!10} 0.9986
    & \cellcolor{gray!10} 0.995
    & \cellcolor{gray!10} 240.68
    & \cellcolor{gray!10} 4.0\% \\
\hline
\multirow{2}{*}{\shortstack[l]{LLaDA/\\C4-en}}
    & i.i.d.\ Gumbel   & 0.8260 & 0.415 & 35740.32 & 10.5\% \\
    & \cellcolor{gray!10} SAC-Copula
    & \cellcolor{gray!10} 0.9138
    & \cellcolor{gray!10} 0.670
    & \cellcolor{gray!10} 494.57
    & \cellcolor{gray!10} 3.0\% \\
\hline
\end{tabular}
\caption{Generalization across models and datasets. Tail failure denotes Composite collapse for Dream-7B/ELI5 and PPL >100 for LLaDA/C4-en}
\label{tab:targeted-transfer}
\vspace{-2em}
\end{table}



\FloatBarrier

\subsection{Residual Evidence under Controlled Token Edits}
\label{sec:exp-mild-edits}

\begin{figure}[!b]
\centering
\includegraphics[width=0.67\columnwidth]{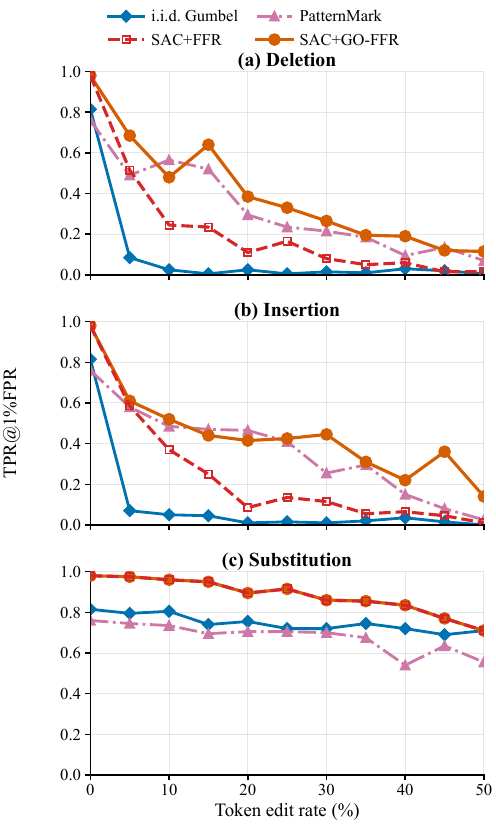}
\caption{TPR at 1\% FPR across controlled 0--50\% token-edit sweeps
for i.i.d.\ Gumbel, PatternMark, SAC+FFR, and SAC+GO-FFR.
GO-FFR provides coarse synchronization recovery under deletion/insertion;
under substitution, the two SAC readouts coincide.}
\label{fig:token-edit-tpr1-camera-ready}
\end{figure}

Figure~\ref{fig:token-edit-tpr1-camera-ready} reports rate-dependent
low-FPR evidence under controlled deletion, insertion, and substitution
for i.i.d.\ Gumbel, PatternMark, SAC+FFR, and SAC+GO-FFR.
Under deletion and insertion, GO-FFR retains more low-FPR evidence than
fixed-alignment FFR once edits are applied, while PatternMark provides
an external cross-method reference. Under substitution, the two SAC
readouts coincide because positional alignment is preserved.
Full AUC/TPR sweeps and additional selected operating-point diagnostics are reported in Appendix~\ref{app:mild-attacks}.

GO-FFR scans a coarse global-offset family, while cumulative
token--tape drift under deletion and insertion remains unresolved.
Semantic/document paraphrasing replaces, reorders, or regenerates the
evidence-bearing sequence and remains outside the current robustness
scope. Full failure diagnostics are in
Appendix~\ref{app:failure-analysis}.

\FloatBarrier

\section{Discussion}
\label{sec:discussion-limitations}

SAC-Copula focuses on smooth, locally correlated DLM Gumbel fields and achieves a strong quality--detectability trade-off. Relative to matched i.i.d.\ Gumbel, its clearest quality gain is improved upper-tail stability with a competitive broader profile. The latent-field analysis is a mechanism-level rationale: one-coordinate marginals neither imply invariance of the complete generated-text distribution nor prove downstream text quality.

FFR provides a SAC-aware covariance-matched readout under the Gaussian mean-shift rationale. GO-FFR extends this readout with coarse global-offset recovery for insertion/deletion.

The size/source controls and frozen-threshold C4 test quantify calibration sensitivity and threshold transfer. LLaDA--ELI5 remains the primary setting, while Dream-7B--ELI5 and LLaDA--C4-en extend the evaluation to a second DLM backbone and a second task/source setting. The fixed-\(W,\sigma_K\) six-point sweep identifies an interior operating region around \(\rho=0.6\).

\section{Limitations}
\label{sec:limitations}

Detection relies on calibrated \(H_0\) samples and development data. Deployment requires private-tape access and representative calibration data. Natural extensions include broader architectural coverage, more systematic \(W/\sigma_K\) studies, runtime characterization, and deployment-oriented robustness evaluation~\citep{P32_sok_watermarking_ai_generated_content}.

Under edit stress tests, GO-FFR partially recovers coarse insertion/deletion offsets, while low-FPR detection degrades as cumulative drift grows. Paraphrastic rewriting remains challenging because it breaks the fixed tape readout~\citep{P26_paraphrasing_evades_detectors,P15_semstamp}. Broader watermark-security work studies strong-watermark impossibility limits and watermark stealing~\citep{P33_watermarks_in_the_sand,P31_watermark_stealing}.

\bibliography{custom}

\appendix
\section{Watermark Properties}
\label{app:watermark-properties}

This appendix gives proof details behind the SAC-Copula field in Section~\ref{sec:method-sac}.  The claims are about the latent watermark tape.  They are not claims about invariance of the complete output-text law.

\subsection{Boundary Convention}
\label{app:watermark-boundary}

The current convolution convention uses zero padding and divides by the global full-window energy
\begin{equation}
S_2=\sum_{k=-h}^{h}K_W[k]^2.
\end{equation}
Therefore exact unit variance is guaranteed for full-window interior positions.  Boundary positions can have smaller variance because fewer nonzero taps contribute.

An all-position idealization would define
\begin{align}
\mathcal K_t
&=
\{k\in[-h,h]:
1\le t-k\le L_{\max}\},
\\
S_2(t)
&=
\sum_{k\in\mathcal K_t}K_W[k]^2,
\end{align}
and divide the boundary convolution by \(\sqrt{S_2(t)}\).  In this paper, strict marginal claims are made under the full-window or boundary-normalized convention.  Boundary deviations under zero padding are treated as implementation checks rather than as part of the exact marginal claim.

\subsection{Marginal Gaussian and Gumbel Exactness}
\label{app:watermark-marginal}

At a full-window position,
\begin{equation}
Z_{\mathrm{spatial}}(t,j)
=
\sum_{k=-h}^{h}
K_W[k]Z_{\mathrm{white},1}(t-k,j).
\end{equation}
This is a linear combination of independent Gaussian variables, hence Gaussian.  Since the source variables have unit variance,
\begin{equation}
\operatorname{Var}
\!\left(Z_{\mathrm{spatial}}(t,j)\right)
=
\sum_{k=-h}^{h}K_W[k]^2
=S_2.
\end{equation}
Thus
\begin{equation}
\widetilde Z_{\mathrm{spatial}}(t,j)
=
\frac{Z_{\mathrm{spatial}}(t,j)}{\sqrt{S_2}}
\sim \mathcal N(0,1).
\end{equation}
The same proof applies to the boundary-normalized ideal form after replacing \(S_2\) by \(S_2(t)\).

The white stream and spatial stream are independent.  Therefore
\begin{equation}
\begin{aligned}
Z_{\mathrm{final}}(t,j)
&=
\sqrt{1-\rho^2}Z_{\mathrm{white},2}(t,j)
\\
&\quad+
\rho\widetilde Z_{\mathrm{spatial}}(t,j)
\end{aligned}
\end{equation}
is Gaussian with zero mean and variance \((1-\rho^2)+\rho^2=1\).  Hence each valid coordinate of \(Z_{\mathrm{final}}\) is \(\mathcal N(0,1)\).

By the probability integral transform,
\begin{equation}
U(t,j)=\Phi(Z_{\mathrm{final}}(t,j))
\sim \operatorname{Uniform}(0,1).
\end{equation}
Applying the inverse standard Gumbel CDF gives
\begin{equation}
G_{\mathrm{corr}}(t,j)
=
F_G^{-1}(U(t,j))
\sim \operatorname{Gumbel}(0,1).
\end{equation}
For the standard Gumbel distribution, \(F_G^{-1}(u)=-\log(-\log u)\).
In numerical evaluation, \(U\) is clamped to avoid floating-point extremes, so exact unbounded Gumbel statements refer to the ideal transform.

\subsection{Finite-Window Latent Autocorrelation}
\label{app:watermark-autocorr}

Let
\begin{equation}
b_k=\frac{K_W[k]}{\sqrt{S_2}},
\qquad
b_k=0\quad\text{for }k\notin[-h,h].
\end{equation}
Then
\begin{equation}
\widetilde Z_{\mathrm{spatial}}(t,j)
=
\sum_k b_k Z_{\mathrm{white},1}(t-k,j).
\end{equation}
Expanding the covariance and using independence of the white source gives the finite-window autocorrelation
\begin{equation}
r_\tau(W,\sigma_K)
=
\sum_k b_k b_{k+\tau}.
\end{equation}
Thus \(r_0=1\), \(r_{-\tau}=r_\tau\), and \(r_\tau=0\) for \(|\tau|>W-1\).  For the final latent field and \(\tau\ne0\),
\begin{equation}
\begin{split}
&\operatorname{Cov}\!\left(
Z_{\mathrm{final}}(t,j),
Z_{\mathrm{final}}(t+\tau,j)
\right)
\\
&\quad =
\rho^2 r_\tau(W,\sigma_K),
\end{split}
\end{equation}
because the white stream contributes no cross-position covariance.  This latent autocorrelation motivates the detector-side \(H_\omega\) template, while the detector covariance itself is estimated from extracted Native evidence.

For the lag-one local-difference proxy used in Section~\ref{sec:why-sac-smooth}, the unit marginal variance and the covariance above give
\begin{equation}
\begin{aligned}
D_{\mathrm{SAC}}
&=
\mathbb E\!\left[
\left(
Z_{\mathrm{final}}(t+1,j)-Z_{\mathrm{final}}(t,j)
\right)^2
\right]
\\
&=
2-2\rho^2 r_1(W,\sigma_K)
=
2(1-\rho^2r_1).
\end{aligned}
\end{equation}
Thus the Section~\ref{sec:why-sac-smooth} expression is a direct consequence of the finite-window autocorrelation and the preserved one-coordinate latent variance.

\subsection{Latent Tape KL Proxy}
\label{app:watermark-kl}

Let \(P_0=\mathcal N(0,I_T)\) be an i.i.d. latent Gaussian tape and \(P_\rho=\mathcal N(0,\Sigma_T)\) the SAC latent Gaussian tape, where
\begin{align}
\Sigma_T
&=
(1-\rho^2)I_T+\rho^2R_T,
\\
[R_T]_{ab}
&=
r_{a-b}(W,\sigma_K).
\end{align}
Assume \(\Sigma_T\succ0\), so the log-determinant is well-defined.  This holds for \(\rho<1\) when \(R_T\) is positive semidefinite; the endpoint \(\rho=1\) additionally requires nonsingularity of \(R_T\).

For zero-mean Gaussians,
\begin{equation}
D_{\mathrm{KL}}(P_\rho\parallel P_0)
=
\frac12
\left(
\operatorname{tr}\Sigma_T
-T
-\log\det\Sigma_T
\right).
\end{equation}
Since \(\operatorname{diag}(\Sigma_T)=1\), \(\operatorname{tr}\Sigma_T=T\), and
\begin{equation}
\mathcal L_{\mathrm{tape}}^{(T)}
=
\frac{1}{T}
D_{\mathrm{KL}}(P_\rho\parallel P_0)
=
-\frac{1}{2T}\log\det\Sigma_T.
\end{equation}
This is a latent process diagnostic, not semantic KL and not output-text KL.

This expansion is an explanatory approximation obtained from the log-determinant series around the identity covariance.  Set \(A=R_T-I_T\).  Since \(\Sigma_T=I_T+\rho^2A\) and \(\operatorname{tr}(A)=0\),
\begin{equation}
\mathcal L_{\mathrm{tape}}^{(T)}
=
\frac{\rho^4}{4T}
\operatorname{tr}(A^2)
+O(\rho^6).
\end{equation}
For the Toeplitz finite-window case,
\begin{equation}
\operatorname{tr}(A^2)
=
\sum_{\tau\ne0}
(T-|\tau|)
r_\tau(W,\sigma_K)^2.
\end{equation}
The leading term is \(O(\rho^4)\) because the diagonal marginal variances are preserved and the trace term cancels.

\subsection{Implementation Checks}
\label{app:watermark-diagnostics}

Implementation checks should include:
\begin{itemize}
    \item marginal Gaussian and Gumbel QQ plots;
    \item empirical versus theoretical autocorrelation;
    \item low-frequency energy of the noise tape;
    \item effective sample size estimates;
    \item boundary variance curves.
\end{itemize}

For a diagnostic sequence \(a\), with discrete Fourier transform \(\widehat a(f)\), define the low-frequency energy fraction as
\begin{equation}
\operatorname{LFE}(a)
=
\frac{
\sum_{f\in\mathcal F_{\mathrm{low}}}
|\widehat a(f)|^2
}{
\sum_{f\in\mathcal F}
|\widehat a(f)|^2
}.
\end{equation}
These checks validate the noise-tape construction, not the distributional invariance of generated text.

\section{Detector Rationale}
\label{app:detector-rationale}

This appendix expands the detector rationale in Section~\ref{sec:method-detection}.  It clarifies why the detector is matched to the evidence geometry induced by the SAC-Copula perturbation field.

\subsection{From i.i.d. Evidence to Covariance-Aware Calibration}
\label{app:detector-covcal}

The old detector aggregates raw evidence with an equal-weight i.i.d. Z-score:
\begin{equation}
Z_{\mathrm{old}}
=
\frac{
\mathbf 1^\top(\mathbf s-\mu_G\mathbf 1)
}{
\sigma_G\sqrt{T}
}.
\end{equation}
It implicitly assumes
\begin{equation}
\Sigma_{x,0}\approx\sigma_G^2I,
\qquad
\Delta_x\propto\mathbf 1.
\end{equation}
For non-i.i.d. evidence, however, the null variance of the equal-weight sum is
\begin{equation}
\operatorname{Var}_0(\mathbf 1^\top\mathbf x)
=
\mathbf 1^\top\Sigma_{x,0}\mathbf 1.
\end{equation}
Positive local dependence increases this variance relative to the i.i.d. value, reducing effective sample size.  HAC or covariance-aware calibration corrects this null variance while keeping the evidence direction \(\mathbf 1\).  If a calibration-only score is a positive rescaling,
\begin{equation}
Z_{\mathrm{h0\_hac}}=c_dZ_{\mathrm{old}},
\qquad c_d>0,
\end{equation}
then within the same detector group and fixed calibration setting it does not change ROC ordering or AUC.  This explains why calibration alone is insufficient when the useful watermark shift is not aligned with \(\mathbf 1\).

\subsection{Why Filtering is Needed for SAC-Copula Evidence}
\label{app:detector-filtering}

SAC-Copula injects a locally correlated, low-lag perturbation field.  The resulting evidence is not merely a uniform mean shift across positions.  Full Filtered Ridge maps centered raw evidence into a SAC-aligned filtered feature:
\begin{equation}
\mathbf u=H_\omega\mathbf x.
\end{equation}
The filter \(H_\omega\) is motivated by the autocorrelation envelope of the SAC kernel, using weights proportional to \(\exp(-\tau^2/(4\sigma_K^2))\) with boundary renormalization.  Concretely, with
\begin{equation}
a_\tau=
\exp\!\left(-\frac{\tau^2}{4\sigma_K^2}\right)
\mathbf 1\{|\tau|\le B\},
\end{equation}
a row-normalized filtered readout can be written as
\begin{equation}
[H_\omega]_{t,s}
=
\frac{
a_{s-t}
}{
\sum_{r:\,1\le t+r\le T}a_r
}.
\end{equation}
where \(a_r=0\) outside \([-B,B]\).
This filter is a detector-side transform, not generation noise.

Native text and the decoding process may also introduce background correlations.  Therefore the latent tape covariance should not be used directly as the detector's null covariance.  The detector estimates the observed-evidence Native covariance and projects it into filtered space:
\begin{equation}
\widehat\Sigma_{u,0}
=
H_\omega
\widehat\Sigma_{x,0}
H_\omega^\top.
\end{equation}
Thus FFR combines a SAC-motivated feature map with Native \(H_0\) observed-evidence covariance whitening.

\subsection{Ridge-Regularized Matched Direction}
\label{app:detector-matched}

In the detector statistic, raw evidence is centered before filtering, so \(\mathbf u=H_\omega\mathbf x\) is already \(H_0\)-centered up to finite-sample error.  The surrogate \(T_w(\mathbf u)=w^\top(\mathbf u-\boldsymbol\mu_{u,0})\) writes the same idea in abstract filtered-space notation.

In filtered feature space, use the Gaussian mean-shift surrogate
\begin{align}
H_0:\quad
\mathbf u
&\sim
(\boldsymbol\mu_{u,0},\Sigma_{u,0}),
\\
H_1:\quad
\mathbf u
&\sim
(\boldsymbol\mu_{u,0}+\boldsymbol\Delta_u,\Sigma_{u,0}).
\end{align}
For a linear score
\begin{equation}
T_w(\mathbf u)
=
w^\top(\mathbf u-\boldsymbol\mu_{u,0}),
\end{equation}
the standardized separation is
\begin{equation}
\operatorname{SNR}(w)
=
\frac{
w^\top\boldsymbol\Delta_u
}{
\sqrt{w^\top\Sigma_{u,0}w}
}.
\end{equation}
Let \(b=\Sigma_{u,0}^{1/2}w\), so \(w=\Sigma_{u,0}^{-1/2}b\), and
\begin{equation}
a=
\Sigma_{u,0}^{-1/2}
\boldsymbol\Delta_u .
\end{equation}
Then
\begin{equation}
\begin{aligned}
\operatorname{SNR}(w)
&=
\frac{
b^\top a
}{
\|b\|_2
}
\\
&\le
\|a\|_2.
\end{aligned}
\end{equation}
Equality holds when \(b\propto a\), giving
\begin{equation}
w^\star
\propto
\Sigma_{u,0}^{-1}\boldsymbol\Delta_u.
\end{equation}

The practical direction is the ridge-regularized plug-in estimate
\begin{equation}
w_{\lambda}
=
(\widehat\Sigma_{u,0}+\lambda I)^{-1}
\widehat{\boldsymbol\Delta}_u.
\end{equation}
Because the main detector centers evidence in raw space before filtering, the implemented scalar score can be written without an additional filtered-space mean subtraction:
\begin{equation}
Z_{\mathrm{FFR}}(y)
=
\frac{
\widehat{\boldsymbol\Delta}_u^\top
(\widehat\Sigma_{u,0}+\lambda I)^{-1}
\mathbf u(y)
}{
\gamma_{\mathrm{FFR}}
}.
\end{equation}
For scores expressed directly in centered raw-evidence space, the corresponding linear direction is
\begin{align}
\mathbf q
&=
\frac{
H_\omega^\top w_\lambda
}{
\gamma_{\mathrm{FFR}}
},
\\
\gamma_{\mathrm{FFR}}
&=
\sqrt{
\widehat{\boldsymbol\Delta}_u^\top
(\widehat\Sigma_{u,0}+\lambda I)^{-1}
\widehat{\boldsymbol\Delta}_u
}.
\end{align}
The ridge is finite-sample regularization for an empirical covariance inverse.  The direction \(\widehat{\boldsymbol\Delta}_u\) must be estimated from a development split or cross-fitting protocol without evaluation leakage.  The optimality statement is limited to the linear detector class under the shared-covariance Gaussian surrogate.

\subsection{Relation to the Original Detector}
\label{app:detector-old-relation}

FFR reduces to the old detector under
\begin{align}
H_\omega&=I,
&
\widehat\Sigma_{u,0}&=\sigma_G^2I,
\\
\widehat{\boldsymbol\Delta}_u&\propto\mathbf 1,
&
\lambda&=0,
\\
\widehat{\boldsymbol\mu}_{x,0}&=0.
\end{align}
In this case the FFR direction is proportional to \(\mathbf 1\), and the score becomes the old equal-weight Gumbel evidence sum.  FFR is therefore a covariance-aware, direction-aware, SAC-aware generalization of the original i.i.d. detector.

\subsection{Global Offset Extension}
\label{app:detector-go}

FFR assumes fixed token-to-tape alignment \(j=i\).  Insertion and deletion can instead induce
\begin{equation}
j=i+\delta(i),
\end{equation}
where \(\delta(i)\) is an unknown drift.  For an alignment \(A\), the ideal alignment-aware centered evidence would be
\begin{equation}
x_i^A
=
s(y_{A(i)};i)
-\mu_G
-\widehat\mu_{x,0,i},
\end{equation}
with score
\begin{equation}
Z(A)=\mathbf q^\top\mathbf x^A.
\end{equation}
GO-FFR restricts alignment to the global-offset family
\begin{equation}
\mathcal A_{\mathrm{GO}}
=
\{A_d(i)=i+d:\ d\in\mathcal D\},
\end{equation}
where \(\mathcal D\subset\mathbb Z\) is a finite candidate offset set.
In the reported attack-aware implementation,
\(\mathcal D=\{-D_{\max},\ldots,0\}\) for deletion,
\(\mathcal D=\{0,\ldots,D_{\max}\}\) for insertion,
and \(\mathcal D=\{0\}\) for fixed-alignment clean or substitution
settings, with \(D_{\max}=96\).
The resulting score is
\begin{align}
Z_{\mathrm{GO\text{-}FFR}}(Y)
&=
\max_{d\in\mathcal D} Z_d(Y),
\\
Z_d(Y)
&=
\mathbf q^\top\mathbf x^{(d)}(Y).
\end{align}
This is a GLRT-style scan over a small alignment family.  The max over offsets inflates the Native \(H_0\) right tail.  The GO-specific threshold is the empirical \(1-\eta\) quantile of \(Z_{\mathrm{GO\text{-}FFR}}\) on Native \(H_0\) samples under the same offset-scan rule.  GO-FFR recovers coarse global synchronization within the scanned global-offset family; cumulative local drift and semantic rewriting remain outside this alignment family.

\subsection{Scope and Limitations}
\label{app:detector-scope}

FFR adapts detection to correlated SAC-Copula evidence, while GO-FFR extends the same scoring direction with coarse synchronization recovery for mild insertion/deletion.  Low-FPR evaluation uses detector-specific Native \(H_0\) calibration, and \(\widehat{\boldsymbol\Delta}_u\) is estimated without evaluation leakage.  Semantic rewriting and cumulative edit-path drift remain outside the global-offset family analyzed here.

\section{Additional Experiment Evidence}
\label{app:experiment-evidence}

This appendix records supporting clean-setting summaries, frozen-output re-analysis, calibration and targeted-transfer controls, and frozen attack-summary evidence used in Section~\ref{sec:experiments}.  The clean quality and clean detection claims in the main text are supported by the frozen clean multi-baseline sweep summaries, not by the attack-summary evidence.  The attack sources are summary-only artifacts derived from token-level edit outputs and semantic/paraphrase attack summaries; the attack-detail and failure-analysis sections do not use per-sample generated text, exact noise tapes, or full detector-score files.

\begin{table*}[t]
\centering
\scriptsize
\setlength{\tabcolsep}{4pt}
\resizebox{\textwidth}{!}{%
\begin{tabular}{lll}
\hline
Evidence item & Appendix location & Source / scope note \\
\hline
Full experimental setup & Tables~\ref{tab:app-generation-watermark-config}--\ref{tab:app-detector-evaluator-config} & implementation scripts, run metadata, and frozen summary files. \\
Metric definitions and orientation & Table~\ref{tab:app-metric-definitions} & Defines raw metrics, direction, and visualization-only quality score. \\
Full clean quality metrics & Tables~\ref{tab:full_quality_metrics}--\ref{tab:full_quality_metrics_diversity} & frozen clean multi-baseline quality summaries. \\
Radar normalization & Tables~\ref{tab:radar_normalization}--\ref{tab:radar_normalized_scores} & Display normalization only; not primary evidence. \\
Clean quality--detection view & Figure~\ref{fig:app_quality_detection_heatmap} & Compact appendix view of clean quality and detection trade-offs. \\
Full clean detection sweep & Figure~\ref{fig:app-clean-detection-sweep} & External-baseline strength sweep supporting the clean multi-method comparison. \\
Unigram strength trade-off & Figure~\ref{fig:app-unigram-tradeoff} & Contrastive static-bias sweep. \\
SAC parameter sensitivity & Table~\ref{tab:sac-parameter-sensitivity} & Raw-metric support for the selected SAC operating point. \\
Detector comparison & Table~\ref{tab:sac-detector-comparison} & Clean old-readout / FFR / GO-FFR sanity check. \\
Token-edit sweeps & Figures~\ref{fig:app-token-edit-sweep-deletion}--\ref{fig:app-token-edit-sweep-substitution} & AUC, TPR@FPR1, and TPR@FPR5 across 0--50\% edit rates. \\
Failure analysis & Tables~\ref{tab:app-failure-edit-drift}--\ref{tab:app-failure-semantic} & Boundary evidence for edit drift and semantic rewriting. \\
Tail stability & Tables~\ref{tab:app-tail-descriptive}--\ref{tab:app-tail-paired} & Frozen-output re-analysis and paired upper-tail diagnostics. \\
Six-point \(\rho\) sweep & Tables~\ref{tab:app-rho-detection}--\ref{tab:app-rho-quality} & Fixed-parameter correlation-strength control. \\
Calibration controls & Tables~\ref{tab:app-cal-size-roc}--\ref{tab:app-cal-prompt-shift} & Size, source, threshold, and prompt-shift diagnostics. \\
Frozen C4 FPR & Table~\ref{tab:app-frozen-c4-fpr} & Detector-only transfer of frozen ELI5 operating thresholds. \\
Dream-7B transfer & Table~\ref{tab:app-dream-transfer} & Targeted second-backbone evidence on ELI5. \\
C4-en transfer & Table~\ref{tab:app-c4-method-matched} & Targeted method-matched continuation evidence. \\
C4 common-detector control & Table~\ref{tab:app-c4-common-detector} & Separate detector-control protocol for C4 generation. \\
Detector derivation & Appendix~\ref{app:detector-rationale} & Full detector definitions are not duplicated in this experiment appendix. \\
\hline
\end{tabular}
}
\caption{Experiment appendix evidence index.  Each row points to the compact appendix evidence used to support Section~\ref{sec:experiments}.  The index is organizational only and does not introduce new experimental values.}
\label{tab:app-evidence-index}
\end{table*}

\subsection{Clean Quality and Detection Details}
\label{app:clean_quality_details}

This subsection records supporting material for the clean multi-baseline comparison in Section~\ref{sec:exp-quality} and Section~\ref{sec:exp-clean-detection}.  The source tables are frozen clean-text sweep summaries used to construct the main trade-off table, quality-score components, radar normalization, and heatmap points.  These tables explain the compact quality score and visualization normalization; they do not introduce additional experimental claims beyond the clean-setting trade-off reported in the main text.  Table~\ref{tab:app-metric-definitions} defines metric directions, Tables~\ref{tab:full_quality_metrics}--\ref{tab:full_quality_metrics_diversity} report the full raw quality metrics, Tables~\ref{tab:radar_normalization}--\ref{tab:radar_normalized_scores} document the radar normalization, and Figures~\ref{fig:app_quality_detection_heatmap}--\ref{fig:app-clean-detection-sweep} provide compact clean quality--detection and clean detection views.
\begin{table*}[t]
\centering
\scriptsize
\setlength{\tabcolsep}{4pt}
\resizebox{\textwidth}{!}{%
\begin{tabular}{llll}
\hline
Metric & Definition / source & Direction & Use in this paper \\
\hline
PPL & Perplexity under the configured evaluator model. & Lower is better & Fluency and tail-stability evidence. \\
PPL-P90 & 90th-percentile conditional PPL. & Lower is better & Radar tail-stability axis. \\
Median / P95 / P99 PPL & Quantiles of the per-output conditional-PPL distribution. & Lower is better & Central and upper-tail summaries. \\
LLaDA PPL\(>100\) & Fraction of evaluated outputs with \(\texttt{Conditional\_PPL}>100\). & Lower is better & LLaDA severe-tail diagnostic. \\
Dream Composite collapse & Preregistered Dream PPL/repetition criterion; distinct from LLaDA PPL\(>100\). & Lower is better & Dream-specific collapse statistic. \\
SBERT & Sentence-embedding similarity to the Native reference. & Higher is better & Semantic preservation. \\
BERTScore & Token-level contextual similarity. & Higher is better & Semantic preservation in full quality table. \\
ROUGE-L & Longest-common-subsequence lexical overlap. & Higher is better & Radar lexical overlap axis. \\
Token-F1 & Native token-set overlap F1. & Higher is better & Radar token-overlap axis. \\
MAUVE & Distributional similarity against Native reference generations. & Higher is better & Distributional fidelity. \\
Token TV / Token JS & Unigram token-distribution divergence from Native. & Lower is better & Token-distribution drift. \\
Top-\(k\) overlap & Overlap with Native top-\(k\) token set. & Higher is better & Token-distribution fidelity. \\
Rep3 & Repeated 3-gram rate. & Lower is better & Repetition diagnostic. \\
Distinct3 & Distinct 3-gram ratio. & Higher is better & Lexical diversity. \\
Ent3 & 3-gram entropy. & Higher is better & Diversity / spread diagnostic. \\
AUC & Threshold-independent detector ranking metric. & Higher is better & Clean and attack diagnostic detection. \\
TPR@FPR1 / TPR@FPR5 & True-positive rate at 1\% / 5\% false-positive rate. & Higher is better & Low-FPR clean detection and attack diagnostics. \\
Quality & Direction-normalized visualization summary. & Higher is better & Descriptive radar/heatmap display only; not primary evidence. \\
\hline
\end{tabular}
}
\caption{Metric definitions and orientations used in Section~\ref{sec:experiments}.  Raw metrics are the primary evidence.  The normalized Quality score is a descriptive display summary constructed from direction-normalized components and should not be interpreted as an additional ground-truth quality label.}
\label{tab:app-metric-definitions}
\end{table*}

\subsection{Full Experimental Configuration}
\label{app:full-experimental-configuration}

This appendix records implementation-level details for reproducibility. Fields not exposed by the frozen scripts or summary metadata are marked as unavailable rather than inferred.

\begin{table*}[t]
\centering
\scriptsize
\setlength{\tabcolsep}{4pt}
\resizebox{\textwidth}{!}{%
\begin{tabular}{lll}
\hline
Category & Field & Setting \\
\hline
Generation & Model name & \texttt{LLaDA-8B-Instruct} \\
Generation & Prompt source & \texttt{ELI5.jsonl} \\
Generation & Prompt format & LLaDA chat template \\
Generation & Split sizes & Native \(H_0=500\), Delta-dev \(=200\), eval \(=200\) \\
Generation & Operating GPU & NVIDIA RTX PRO 6000 \\
Generation & Generation length & 256 tokens \\
Generation & Block length & 32 \\
Generation & Diffusion steps & \(\max(1,\lfloor 0.5 \times\) generation length\(\rfloor)=128\) \\
Generation & Temperature & 1.0 \\
Generation & top-\(p\) / top-\(k\) & not available in the frozen configuration fields \\
Generation & Mask id & 126336 \\
Generation & Seeds and key & fixed random seeds and a fixed private watermark key; exact values are kept in the implementation configuration \\
Generation & Cached artifacts & generated texts and exact perturbation tapes are cached for reproducibility in the implementation artifacts \\
Watermark & Main SAC-Copula & \(W=39\), \(\sigma_K=15.0\), \(\rho=0.6\) \\
Watermark & SAC sensitivity rows & \(W=35,\sigma_K=11.5,\rho=0.7\); \(W=35,\sigma_K=13.5,\rho=0.7\); \(W=39,\sigma_K=15.0,\rho=0.6\) \\
Watermark & Perturbation scale & \(\alpha=1.0\), passed as the generation temperature in the generation call \\
Watermark & Tape length & 256 generated positions in the clean evaluation \\
Watermark & Exact tapes & saved under \texttt{exact\_tapes}; missing-tape reconstruction disabled \\
Baselines & Direct mechanism baseline & i.i.d. position-indexed Gumbel \\
Baselines & External baselines & KGW context-green bias, Unigram static-green bias, PatternMark MarkovPattern ReferenceLite \\
Baselines & External sweep
& \(\delta\in\{1,2,3\}\) for KGW, Unigram, and PatternMark;
\(\gamma=0.25\) for KGW and Unigram \\
Baselines & KGW details & conv-kernel PRF, kernel \([-1]\), context width 1, sumhash seeding, no-bias fallback \\
Baselines & Unigram details & static greenlist bias; unique-token detector is the main detector \\
Baselines & PatternMark details
& MarkovPattern ReferenceLite, \(k=2\), pattern length 4, 10000 null samples \\
\hline
\end{tabular}
}
\caption{Generation, watermark, and baseline configuration for the clean multi-baseline evaluation.  Values are taken from the frozen generation/detection script and collected run metadata.}
\label{tab:app-generation-watermark-config}
\end{table*}

\begin{table*}[t]
\centering
\scriptsize
\setlength{\tabcolsep}{4pt}
\resizebox{\textwidth}{!}{%
\begin{tabular}{lll}
\hline
Category & Field & Setting \\
\hline
Detection & Canonical detector mapping & i.i.d. Gumbel: old z-score; KGW: KGWGreenlistZDetectorV2; Unigram: UnigramUniqueGreenlistDetectorV2; PatternMark: PatternMarkMarkovPatternDetector \\
Detection & SAC detector rows & old z-score, Full Filtered Ridge, Global Offset Full Filtered Ridge \\
Detection & FFR ridge & \(\lambda=10^{-4}\) \\
Detection & Cross-fitting & 2 folds with crossfit seed 20260506 for empirical direction estimation \\
Detection & Native \(H_0\) calibration & detector-specific calibration on held-out Native \(H_0\) samples \\
Detection & Target FPRs & 1\% and 5\% \\
Detection & GO offset setting & max offset 96 and offset penalty 0.0; clean GO-FFR scoring uses offset 0 \\
Detection & Tape device & generation tape on runtime device; detection tape on CPU \\
Quality & PPL model & Llama-3.1-8B-Instruct evaluator checkpoint \\
Quality & PPL collapse threshold & PPL \(>100\) \\
Quality & PPL trim quantile & 0.95 \\
Quality & SBERT model & all-MiniLM-L6-v2 sentence-embedding model \\
Quality & BERTScore model & \texttt{roberta-large}, English, no baseline rescaling \\
Quality & MAUVE setup & Native reference group, \texttt{gpt2-large} features, max text length 256, 10 buckets \\
Quality & Token drift & unigram JS, TV, and top-\(k\) overlap for \(k\in\{50,100,500,1000\}\) \\
Quality & Repetition/diversity & Rep3, Distinct3, Ent3, with Self-BLEU 3/4 available in multidimensional quality metadata \\
Quality & Normalized quality score & descriptive direction-normalized summary for visualization, not a primary metric \\
\hline
\end{tabular}
}
\caption{Detector, calibration, and evaluator configuration for the clean multi-baseline evaluation.  Values are taken from the implementation scripts, collected run metadata, and existing summary tables.}
\label{tab:app-detector-evaluator-config}
\end{table*}

\paragraph{Baseline adaptation notes.}
KGW is adapted as a context-greenlist bias baseline with the configured context width, PRF, and detector listed in Table~\ref{tab:app-generation-watermark-config} and Table~\ref{tab:app-detector-evaluator-config}.  Unigram is used as a static-greenlist clean-text trade-off baseline with its unique-token detector.  PatternMark is the MarkovPattern ReferenceLite variant used in the clean evaluation, not a claim of a complete independent reimplementation of every PatternMark setting.  Detector derivations for the old readout, covariance-aware calibration, filtering, ridge matched direction, and GO-FFR are provided in Appendix~\ref{app:detector-covcal}, Appendix~\ref{app:detector-filtering}, Appendix~\ref{app:detector-matched}, and Appendix~\ref{app:detector-go}; this experiment appendix only records how those detectors are used in the experiments.

\begin{table*}[t]
\centering
\scriptsize
\setlength{\tabcolsep}{3pt}
\resizebox{\textwidth}{!}{%
\begin{tabular}{llrrrrrrr}
\hline
Method & Strength & Mean PPL & Median PPL & Collapse & SBERT & BERTScore & MAUVE & Token TV \\
\hline
Native & -- & 16.840 & 6.381 & 0.005 & 1.000 & 1.000 & 1.000 & 0.000 \\
i.i.d. Gumbel & -- & 636.069 & 8.446 & 0.035 & 0.784 & 0.868 & 0.991 & 0.163 \\
KGW & \(\delta=1\) & \(5.32{\times}10^4\) & 6.697 & 0.030 & 0.794 & 0.869 & 0.999 & 0.146 \\
KGW & \(\delta=2\) & \(6.98{\times}10^4\) & 8.020 & 0.045 & 0.776 & 0.867 & 0.998 & 0.173 \\
KGW & \(\delta=3\) & \(8.30{\times}10^4\) & 10.543 & 0.050 & 0.762 & 0.863 & 0.982 & 0.228 \\
Unigram & \(\delta=1\) & 7.714 & 6.145 & 0.000 & 0.803 & 0.867 & 0.981 & 0.233 \\
Unigram & \(\delta=2\) & 7.630 & 6.535 & 0.000 & 0.755 & 0.851 & 0.841 & 0.360 \\
Unigram & \(\delta=3\) & 9.464 & 6.585 & 0.005 & 0.690 & 0.836 & 0.601 & 0.439 \\
PatternMark & \(\delta=1\) & 3883.179 & 6.766 & 0.010 & 0.781 & 0.869 & 0.983 & 0.153 \\
PatternMark & \(\delta=2\) & 3898.094 & 7.263 & 0.015 & 0.779 & 0.866 & 0.996 & 0.161 \\
PatternMark & \(\delta=3\) & 3901.117 & 8.938 & 0.020 & 0.768 & 0.863 & 0.982 & 0.198 \\
SAC-Copula & main & 12.164 & 8.260 & 0.010 & 0.798 & 0.868 & 0.999 & 0.163 \\
\hline
\end{tabular}
}
\caption{Full automatic quality metrics for the clean multibase comparison, part 1.  The table reports fluency and stability, semantic preservation, distributional fidelity, and token-distribution drift.  Lower values are better for PPL, collapse, and token TV; higher values are better for semantic similarity and MAUVE.}
\label{tab:full_quality_metrics}
\end{table*}

\begin{table*}[t]
\centering
\scriptsize
\setlength{\tabcolsep}{3.5pt}
\resizebox{\textwidth}{!}{%
\begin{tabular}{llrrrrrrr}
\hline
Method & Strength & Token JS & Top50 & Top100 & Rep3 & Distinct3 & Ent3 & Quality \\
\hline
Native & -- & 0.000 & 1.000 & 1.000 & 0.406 & 0.594 & 3.796 & 0.960 \\
i.i.d. Gumbel & -- & 0.059 & 0.940 & 0.830 & 0.415 & 0.585 & 3.680 & 0.661 \\
KGW & \(\delta=1\) & 0.057 & 0.920 & 0.900 & 0.414 & 0.586 & 3.798 & 0.694 \\
KGW & \(\delta=2\) & 0.060 & 0.900 & 0.850 & 0.450 & 0.550 & 3.555 & 0.553 \\
KGW & \(\delta=3\) & 0.068 & 0.860 & 0.830 & 0.510 & 0.489 & 3.179 & 0.379 \\
Unigram & \(\delta=1\) & 0.074 & 0.860 & 0.760 & 0.395 & 0.605 & 3.967 & 0.798 \\
Unigram & \(\delta=2\) & 0.120 & 0.660 & 0.600 & 0.455 & 0.545 & 3.819 & 0.545 \\
Unigram & \(\delta=3\) & 0.164 & 0.600 & 0.430 & 0.560 & 0.440 & 3.408 & 0.150 \\
PatternMark & \(\delta=1\) & 0.058 & 0.940 & 0.840 & 0.426 & 0.574 & 3.686 & 0.721 \\
PatternMark & \(\delta=2\) & 0.061 & 0.900 & 0.800 & 0.422 & 0.578 & 3.809 & 0.714 \\
PatternMark & \(\delta=3\) & 0.066 & 0.880 & 0.810 & 0.480 & 0.520 & 3.442 & 0.555 \\
SAC-Copula & main & 0.061 & 0.920 & 0.810 & 0.414 & 0.586 & 3.745 & 0.758 \\
\hline
\end{tabular}
}
\caption{Full automatic quality metrics for the clean multibase comparison, part 2.  Lower values are better for token JS and repetition; higher values are better for vocabulary-overlap, distinctness, entropy, and the normalized quality score.}
\label{tab:full_quality_metrics_diversity}
\end{table*}

\begin{table*}[t]
\centering
\scriptsize
\setlength{\tabcolsep}{4pt}
\resizebox{\textwidth}{!}{%
\begin{tabular}{llll}
\hline
Radar axis & Raw metric source & Original direction & Normalized score meaning \\
\hline
SBERT & SBERT & Higher is better & Embedding-level semantic closeness to Native text. \\
ROUGE-L & Mean ROUGE-L vs.\ Native & Higher is better & Lexical-sequence overlap with Native generations. \\
Token-F1 & Native token-set F1 & Higher is better & Native token-set overlap. \\
Rep3 & Mean Rep3 & Lower is better & Lower repeated 3-gram rate after inverted min--max normalization. \\
Collapse\(>100\) & CollapseRate\_GT100 & Lower is better & Fewer severe PPL-tail collapse cases after inverted min--max normalization. \\
PPL-P90 & Conditional PPL P90 & Lower is better & Lower 90th-percentile conditional perplexity after inverted min--max normalization. \\
\hline
\end{tabular}
}
\caption{Metric orientation used in the v3.8 quality radar.  All radar axes are direction-normalized so that larger values indicate better or more Native-like behavior.  Lower-is-better metrics, including repetition, collapse rate, and PPL-P90, are inverted before normalization.  The normalization is used only for visualization; raw quality metrics remain the primary evidence in Table~\ref{tab:quality-detection}, Table~\ref{tab:full_quality_metrics}, and Table~\ref{tab:full_quality_metrics_diversity}.}
\label{tab:radar_normalization}
\end{table*}

\begin{table*}[t]
\centering
\scriptsize
\setlength{\tabcolsep}{4pt}
\begin{tabular}{lrrrrrr}
\hline
Method & SBERT & ROUGE-L & Token-F1 & Rep3 & Collapse\(>100\) & PPL-P90 \\
\hline
Native & 1.000 & 1.000 & 1.000 & 1.000 & 1.000 & 1.000 \\
i.i.d. Gumbel & 0.869 & 0.948 & 0.767 & 0.992 & 0.333 & 0.830 \\
KGW \(\delta=3\) & 0.666 & 0.734 & 0.562 & 0.340 & 0.080 & 0.080 \\
PatternMark \(\delta=3\) & 0.721 & 0.764 & 0.677 & 0.549 & 0.667 & 0.788 \\
Unigram \(\delta=3\) & 0.080 & 0.080 & 0.080 & 0.080 & 1.000 & 1.000 \\
SAC-Copula & 1.000 & 1.000 & 1.000 & 1.000 & 0.889 & 0.919 \\
\hline
\end{tabular}
\caption{Display-normalized scores used in the v3.8 quality radar.  A small display floor is applied to low-scoring axes to avoid degenerate polygons in the visualization.  These scores are descriptive display values, not additional primary quality metrics.}
\label{tab:radar_normalized_scores}
\end{table*}

\begin{figure*}[t]
\centering
\includegraphics[width=0.95\textwidth]{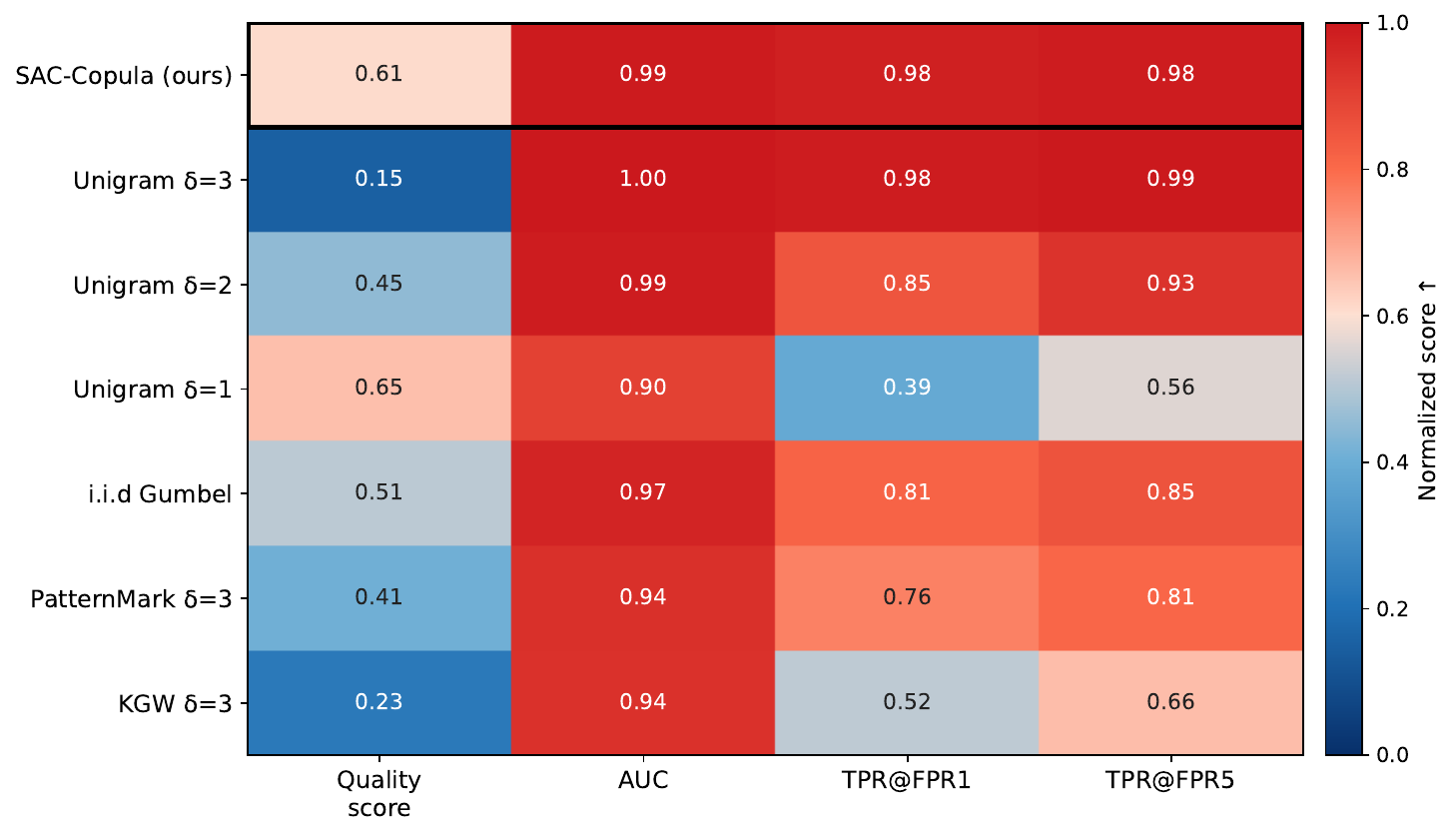}
\caption{Quality--detection heatmap for the clean multibase comparison.  Rows correspond to representative watermark settings, and columns show the normalized quality-preservation score, AUC, TPR@FPR1, and TPR@FPR5.  The heatmap quality score averages semantic preservation, MAUVE, token-distribution fidelity, low repetition, lexical diversity, and tail stability after direction-aware normalization.  SAC-Copula maintains a favorable balance between clean detectability and quality preservation compared with high-strength external baselines.}
\label{fig:app_quality_detection_heatmap}
\end{figure*}

\paragraph{Quality--detection heatmap.}
Figure~\ref{fig:app_quality_detection_heatmap} summarizes the same clean-setting trends using a compact heatmap.  The quality score averages direction-normalized components for semantic preservation, MAUVE-based distributional fidelity, token-distribution fidelity, low repetition, lexical diversity, and PPL-tail stability.  This visualization is intended as a compact summary of the quality--detection balance rather than a replacement for the raw metric values in Table~\ref{tab:full_quality_metrics}.

The external-baseline clean detection sweep over \(\delta\in\{1,2,3\}\) is fully visualized below in Figure~\ref{fig:app-clean-detection-sweep} and supports the clean multi-method comparison in Table~\ref{tab:quality-detection}.  Figure~\ref{fig:app_quality_detection_heatmap} and Table~\ref{tab:sac-detector-comparison} further connect the clean delta sweep to the raw quality tables and detector-readout sanity checks without duplicating the full CSV rows.

\paragraph{Full clean detection sweep.}
Figure~\ref{fig:app-clean-detection-sweep} reports the full clean detection visualization for AUC, TPR@FPR1, and TPR@FPR5 across the external-baseline strength sweep.  This appendix figure provides the full clean-detection sweep across methods and watermark-strength settings and does not introduce new experimental values.

\begin{figure*}[t]
\centering
\includegraphics[width=0.95\textwidth]{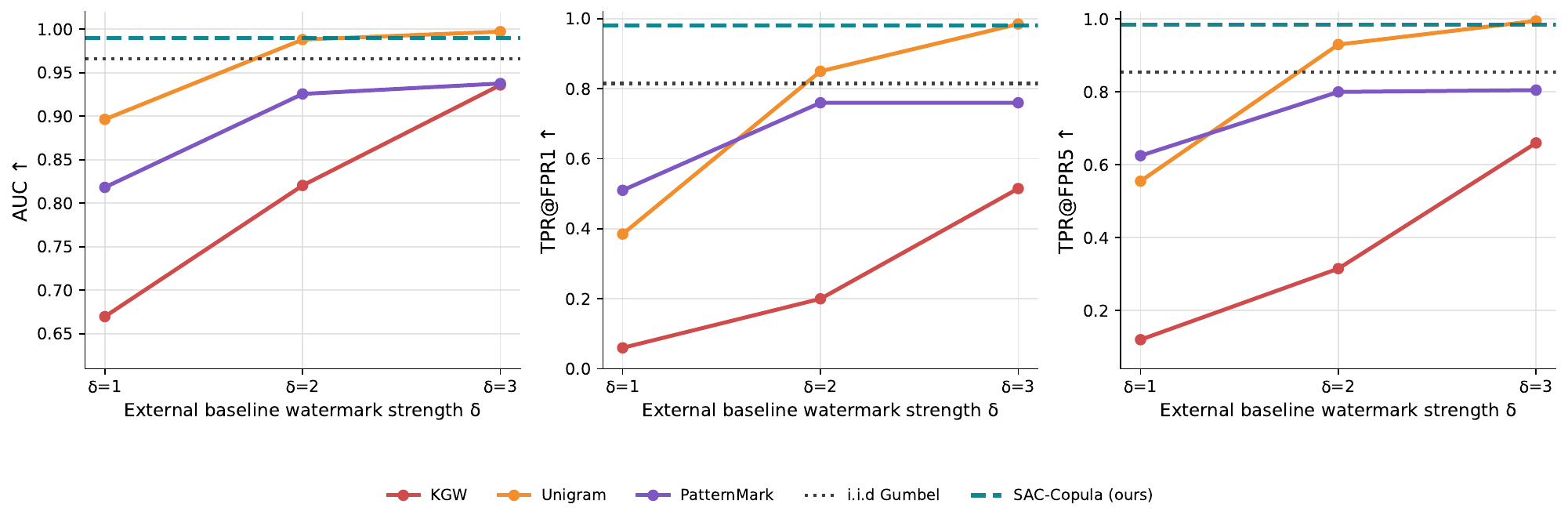}
\caption{Full clean detection sweep across watermark strengths.  The panels report AUC, TPR@FPR1, and TPR@FPR5 for KGW, Unigram, and PatternMark over \(\delta\in\{1,2,3\}\), with i.i.d.\ Gumbel and SAC-Copula shown as reference operating points.  This figure reports clean fixed-alignment detection only.}
\label{fig:app-clean-detection-sweep}
\end{figure*}

\subsection{Unigram Strength Trade-off}
\label{app:unigram-tradeoff}

\begin{figure*}[t]
\centering
\includegraphics[width=0.82\textwidth]{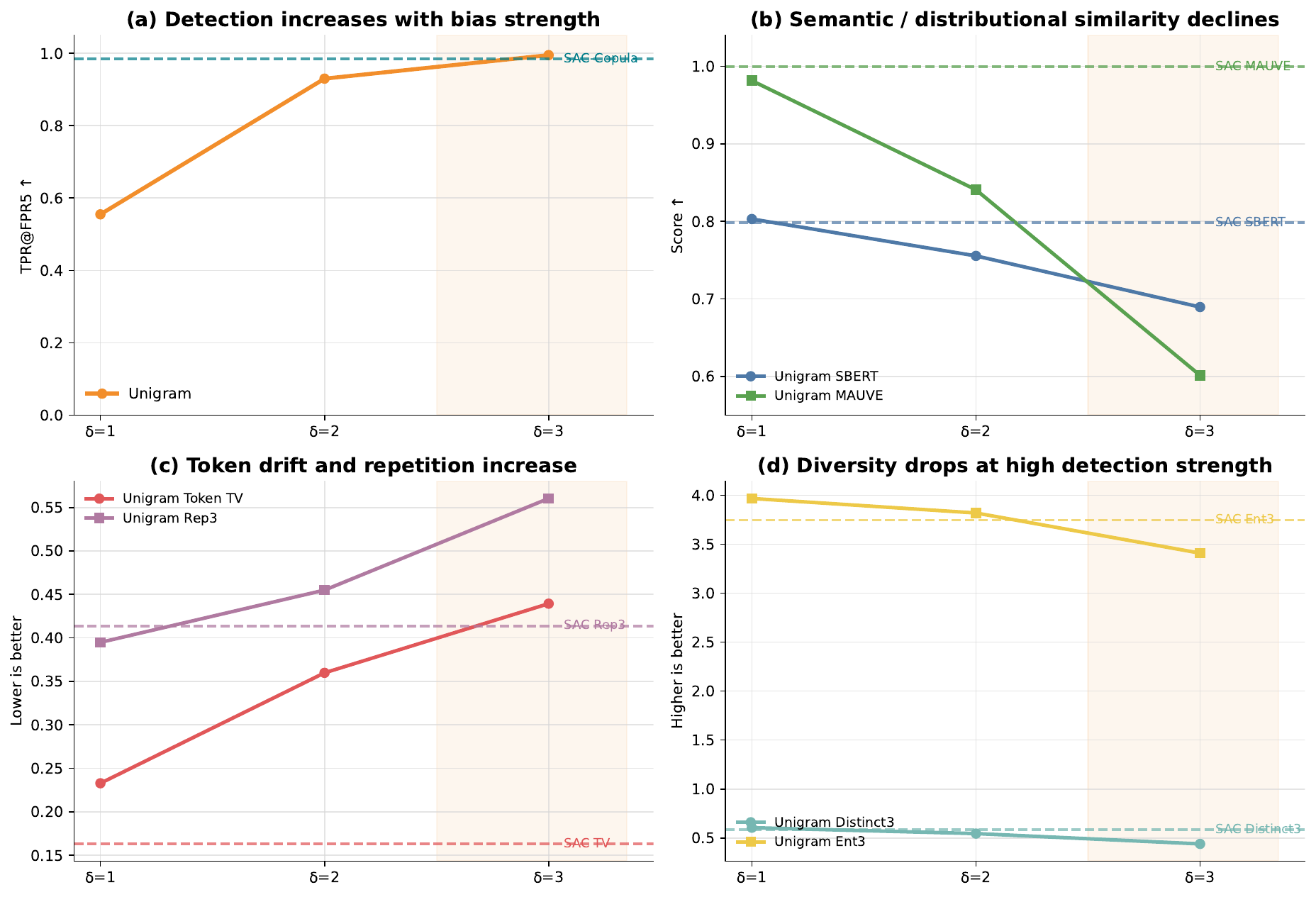}
\caption{SAC-Copula avoids the high-strength static-bias quality trade-off.  The Unigram strength sweep is included as a contrastive reference: Panel (a) shows clean TPR@FPR5 increasing with Unigram strength \(\delta\), while Panels (b)--(d) show the accompanying semantic, distributional, token-drift, repetition, and diversity costs.  SAC-Copula is shown as the reference operating point.}
\label{fig:app-unigram-tradeoff}
\end{figure*}

Figure~\ref{fig:app-unigram-tradeoff} keeps the Unigram strength sweep as a contrastive reference for static-bias tuning.  It is included in the appendix so that the main text can focus on the SAC-Copula clean quality and clean detection profile.  The figure shows that stronger Unigram settings can improve clean detectability, but the same sweep changes semantic, distributional, token-drift, repetition, and diversity dimensions.  This supports the main-text interpretation of Unigram as a trade-off baseline, not as the central object of study.

\subsection{Traceable Clean-Generation Example}
\label{app:quality-examples}

Table~\ref{tab:app-quality-example} provides one traceable qualitative example from the frozen generated-text JSONL artifact.  The example is included only to make the automatic quality metrics more concrete; it is not used as primary evidence and is not selected to prove a general qualitative claim.

\begin{table*}[t]
\centering
\scriptsize
\setlength{\tabcolsep}{4pt}
\resizebox{\textwidth}{!}{%
\begin{tabular}{lll}
\hline
System & Output excerpt \\ \\
\hline
Native & \texttt{Eval|Native|701|256} & Dollar stores turn a profit primarily selling items at prices of only a dollar each. This allows them to sell a large quantity of items at a... \\
i.i.d.\ Gumbel & \texttt{Eval|i.i.d\_Baseline|701|256} & Dollar stores turn a profit primarily by selling low-cost items at a lower price point. These items are usually bulk from larger retailers a... \\
SAC-Copula & \texttt{Eval|SAC\_W39\_S15.0\_R0.6|701|256} & Dollar stores turn a profit through selling a wide variety of items at very low prices. They can offer a wide selection of items in from foo... \\
KGW & \texttt{Eval|KGW\_ContextGreen\_Bias|701|256} & Dollar stores turn a profit by by offering items at discounted prices, which allows them to sell more items per dollar spent. This is done b... \\
Unigram & \texttt{Eval|Unigram\_StaticGreen\_Bias|701|256} & Dollar stores turn a profit by selling a wide range of goods at low prices. They compete by offering a diverse selection of goods, including... \\
PatternMark ReferenceLite & \texttt{Eval|PatternMark\_MarkovPattern\_ReferenceLite|701|256} & Dollar stores make a profit through a combination of factors. 1. Lower pricing: Dollar stores sell items at very low prices, often as little... \\
\hline
\end{tabular}
}
\caption{Traceable clean-generation example for the prompt ``How do dollar stores turn a profit?''  Excerpts are copied from the frozen clean-generation artifact and truncated for space.  The table is illustrative only and does not replace the automatic quality metrics.}
\label{tab:app-quality-example}
\end{table*}

\subsection{SAC Parameter Sensitivity}
\label{app:sac-parameter-sensitivity}

This subsection documents the SAC operating point used in the main experiments.  The three SAC parameters have distinct roles: \(W\) controls the correlation window or range, \(\sigma_K\) controls the kernel decay and smoothness, and \(\rho\) controls the strength of the correlated branch.  Table~\ref{tab:sac-parameter-sensitivity} compares several SAC configurations under the same clean-text evaluation slice.

Detection metrics come from the FFR clean-detection summary, and quality metrics come from the multidimensional quality and token-drift summaries in the shared clean-text evaluation slice indexed as \(\delta=1\) for the external-baseline strength sweep.  This \(\delta\) index is not a SAC parameter; the SAC parameters are \(W,\sigma_K,\rho\).  The normalized quality score is not included in these delta-sweep summary files, so it remains marked as unavailable rather than reconstructed, and the interpretation should rely on the raw listed metrics.

\begin{figure*}[t]
\centering
\includegraphics[width=0.95\textwidth]{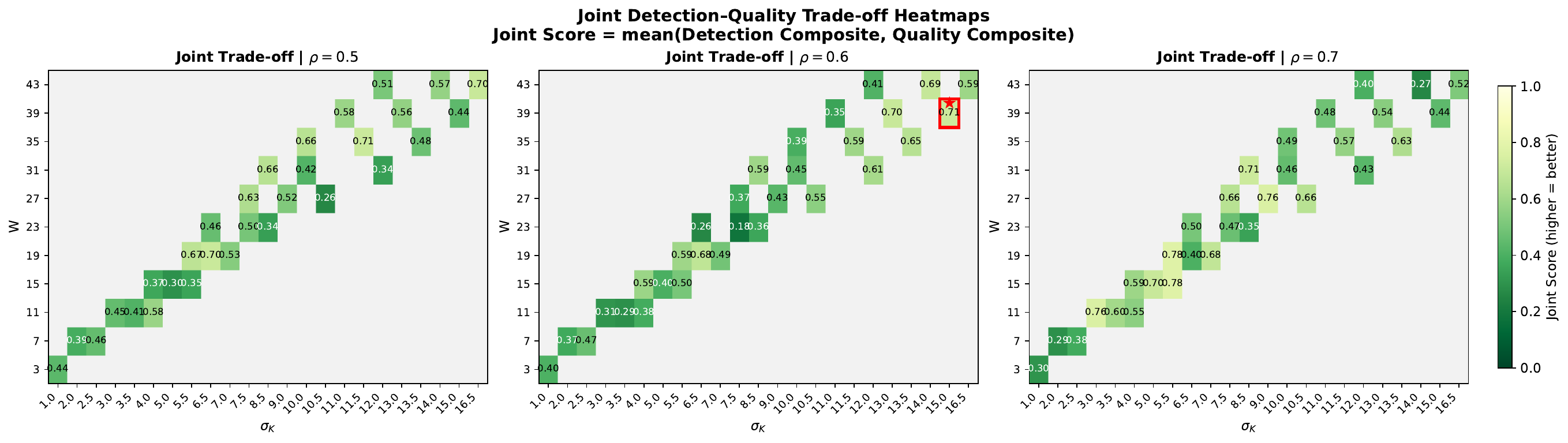}
\caption{SAC parameter sensitivity for clean operating-point selection.  The heatmap groups configurations by \(\rho\), with \(W\) and \(\sigma_K\) on the axes and color indicating the joint quality--detectability score.  The selected setting \(W=39,\sigma_K=15.0,\rho=0.6\) is used as the main SAC-Copula configuration.  This visualization guides operating-point selection and is not a claim of global optimality or raw-metric dominance.}
\label{fig:app-sac-parameter-sensitivity}
\end{figure*}

\begin{table*}[t]
\centering
\scriptsize
\setlength{\tabcolsep}{3pt}
\resizebox{\textwidth}{!}{%
\begin{tabular}{lrrrrrrrrrrrrr}
\hline
Configuration & \(W\) & \(\sigma_K\) & \(\rho\) & AUC & TPR@1\% & TPR@5\% & Mean PPL & Collapse & Quality & MAUVE & Rep3 & Distinct3 & Token TV \\
\hline
SAC W35 S11.5 R0.7 & 35 & 11.5 & 0.7 & 0.984 & 0.895 & 0.965 & 13.240 & 0.005 & -- & 0.998 & 0.443 & 0.557 & 0.184 \\
SAC W35 S13.5 R0.7 & 35 & 13.5 & 0.7 & 0.991 & 0.950 & 0.970 & 11.934 & 0.000 & -- & 0.980 & 0.446 & 0.554 & 0.189 \\
\rowcolor{gray!10}
Selected SAC-Copula & 39 & 15.0 & 0.6 & 0.990 & 0.980 & 0.985 & 12.163 & 0.010 & -- & 0.999 & 0.414 & 0.586 & 0.163 \\
\hline
\end{tabular}
}
\caption{Compact appendix summary of SAC parameter sensitivity under clean DLM generation.  The selected SAC-Copula row corresponds to the main-text configuration \(W=39,\sigma_K=15.0,\rho=0.6\).  The normalized composite Quality score is unavailable in the delta-sweep summary CSVs used for this parameter table, so the Quality column is left blank and the table should be interpreted through the listed raw metrics.  The \(\delta=1\) data slice is a shared clean-text evaluation slice, not a SAC parameter.}
\label{tab:sac-parameter-sensitivity}
\end{table*}

The table should be read as a compact trade-off summary.  The \(W=35,\rho=0.7\) variants can improve some central fluency or collapse indicators, especially for \(W=35,\sigma_K=13.5,\rho=0.7\), but they also shift diversity, token-drift, and low-FPR detection behavior.  The selected \(W=39,\sigma_K=15.0,\rho=0.6\) setting is used because it retains strong low-FPR detection while maintaining favorable MAUVE, repetition/diversity, and token-drift behavior.

Table~\ref{tab:sac-parameter-sensitivity} supports the selected configuration as a deliberate quality--detectability operating point over the reported local sensitivity configurations.

\subsection{Detector Comparison for SAC Evidence}
\label{app:detector-comparison}

This subsection is included to separate generation-side structure from detector-side readout in the clean setting.  The old detector is the i.i.d. equal-weight baseline readout, while FFR is the SAC-aware filtered statistic used to read the local low-lag evidence geometry induced by SAC-Copula.  GO-FFR provides continuity with the mild insertion/deletion stress tests.

Table~\ref{tab:sac-detector-comparison} reports the available clean detector comparison rows from the clean detection summary.  The comparison should be read in two steps: if SAC evidence already improves over i.i.d. Gumbel under the old detector, that supports a generation-side signal difference; if FFR improves further on SAC evidence, that supports a matched readout for SAC structure.

\begin{table*}[t]
\centering
\scriptsize
\setlength{\tabcolsep}{4pt}
\resizebox{\textwidth}{!}{%
\begin{tabular}{llrrrl}
\hline
Method / group & Detector & AUC & TPR@1\% & TPR@5\% & Role \\
\hline
i.i.d. Gumbel & old z-score & 0.966 & 0.815 & 0.855 & Legacy baseline detector. \\
SAC-Copula & old z-score & 0.987 & 0.915 & 0.950 & Sanity check with the old readout. \\
SAC-Copula & FFR & 0.990 & 0.980 & 0.985 & Main SAC-aware clean readout. \\
SAC-Copula & GO-FFR & 0.990 & 0.980 & 0.985 & Offset-enabled variant; same clean score here. \\
SAC W35 S11.5 R0.7 & FFR & 0.984 & 0.895 & 0.965 & Parameter-sweep SAC readout. \\
SAC W35 S13.5 R0.7 & FFR & 0.991 & 0.950 & 0.970 & Parameter-sweep SAC readout. \\
\hline
\end{tabular}
}
\caption{Compact appendix summary of available clean detector comparison rows for i.i.d. Gumbel and SAC-Copula evidence.  FFR and GO-FFR read the local correlated evidence generated by SAC, while the old-detector rows provide sanity checks against the legacy i.i.d. readout.}
\label{tab:sac-detector-comparison}
\end{table*}

The clean fixed-alignment setting explains why GO-FFR matches or closely tracks FFR here: offset recovery is unnecessary when alignment is intact.  Table~\ref{tab:sac-detector-comparison} serves as a clean-detector sanity check connecting the shared legacy readout to the SAC-aware readout; GO-FFR is included for continuity with the controlled edit diagnostics.

\subsection{Controlled Token-Level Edit Details}
\label{app:mild-attacks}

Table~\ref{tab:app-mild-attacks} and Figures~\ref{fig:app-token-edit-sweep-deletion}--\ref{fig:app-token-edit-sweep-substitution} support Section~\ref{sec:exp-mild-edits}.  These results are intended as controlled token-edit diagnostics rather than a claim of adversarial robustness.  Across selected deletion, insertion, and substitution settings, the SAC-aware detector retains partial residual evidence, and the Global Offset variant provides a coarse realignment benefit.  The full sweep figures report AUC, TPR@FPR1, and TPR@FPR5 across 0--50\% attack rates.  Low-FPR TPR remains substantially harder than AUC under deletion and insertion, which supports the failure analysis rather than a broad edit-robustness claim.  These conclusions come from the frozen token-edit summaries, relative-gain, and global-offset summaries.  They should not be inferred from the multi-baseline clean-text results.

\begin{figure*}[t]
\centering
\includegraphics[width=0.95\textwidth]{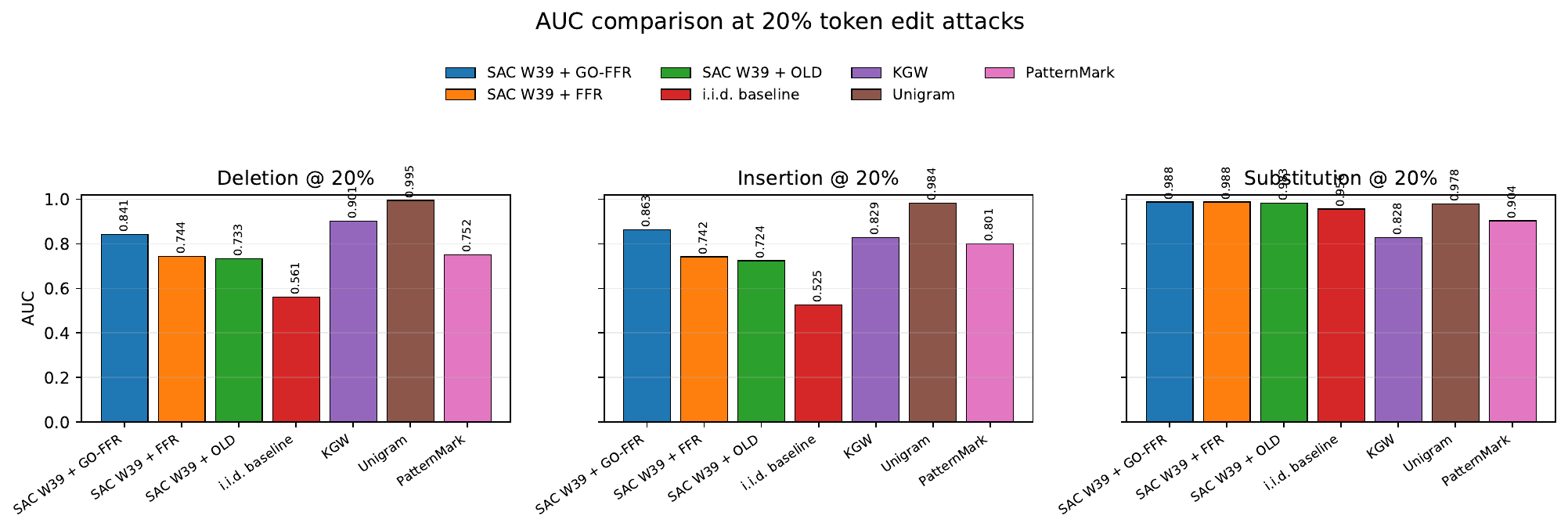}
\caption{Full AUC comparison under \(20\%\) token-level edit attacks.  The three panels report deletion, insertion, and substitution for SAC-Copula readouts, the direct i.i.d.\ Gumbel baseline, and external clean-text baselines.  This full comparison provides the complete cross-method view supporting the compact quantitative token-edit discussion in the main text and is reported as a residual-evidence diagnostic, not as a broad robustness claim.}
\label{fig:app-auc-20pct-token-edit-attacks-full}
\end{figure*}

The full 20\% cross-method AUC and detector/readout comparison is shown in Figure~\ref{fig:app-auc-20pct-token-edit-attacks-full}.  Table~\ref{tab:app-mild-attacks} focuses on the detector-readout rows used to interpret SAC OLD, FFR, and GO-FFR behavior.  Substitution does not induce cumulative position drift. Under substitution, the GO-FFR scan reduces to \(\mathcal D=\{0\}\) and therefore coincides with fixed-alignment FFR; both curves are retained in the sweep figures for consistent cross-panel comparison.

\begin{table*}[t]
\centering
\scriptsize
\setlength{\tabcolsep}{4pt}
\resizebox{\textwidth}{!}{%
\begin{tabular}{llrrrrl}
\hline
Edit & Detector row & AUC & TPR@1\% & TPR@5\% & Best offset & Reading \\
\hline
Deletion 20\% & i.i.d. old & 0.561 & 0.025 & 0.110 & 0.000 & Alignment-changing edit strongly weakens i.i.d. evidence. \\
Deletion 20\% & SAC FFR & 0.780 & 0.150 & 0.370 & 0.000 & SAC-aware filtering retains partial residual evidence. \\
Deletion 20\% & SAC GO-FFR & 0.892 & 0.460 & 0.590 & -27.295 & Offset scanning partially recovers coarse shift. \\
Insertion 20\% & i.i.d. old & 0.525 & 0.010 & 0.045 & 0.000 & Insertions disrupt fixed token-to-tape indexing. \\
Insertion 20\% & SAC FFR & 0.842 & 0.195 & 0.435 & 0.000 & SAC-aware readout remains informative under controlled insertion. \\
Insertion 20\% & SAC GO-FFR & 0.884 & 0.505 & 0.670 & 18.230 & Offset scanning gives a coarse synchronization gain. \\
Substitution 20\% & i.i.d. old & 0.956 & 0.755 & 0.825 & 0.000 & Substitution is less damaging because sequence alignment is preserved. \\
Substitution 20\% & SAC FFR & 0.989 & 0.890 & 0.975 & 0.000 & SAC evidence remains detectable in the indexed substitution setting. \\
\hline
\end{tabular}
}
\caption{Selected low-rate token-level edit diagnostics from the frozen token-edit summary.  These results are reported as partial residual-evidence diagnostics rather than broad robustness claims; the table is not evidence of full edit-path alignment or semantic rewrite robustness.}
\label{tab:app-mild-attacks}
\end{table*}
\clearpage
\begingroup
\makeatletter
\setlength{\@dblfptop}{0pt}
\setlength{\@dblfpsep}{10pt}
\setlength{\@dblfpbot}{0pt plus 1fil}
\makeatother

\begin{figure*}[t]
\centering
\includegraphics[width=0.95\textwidth]{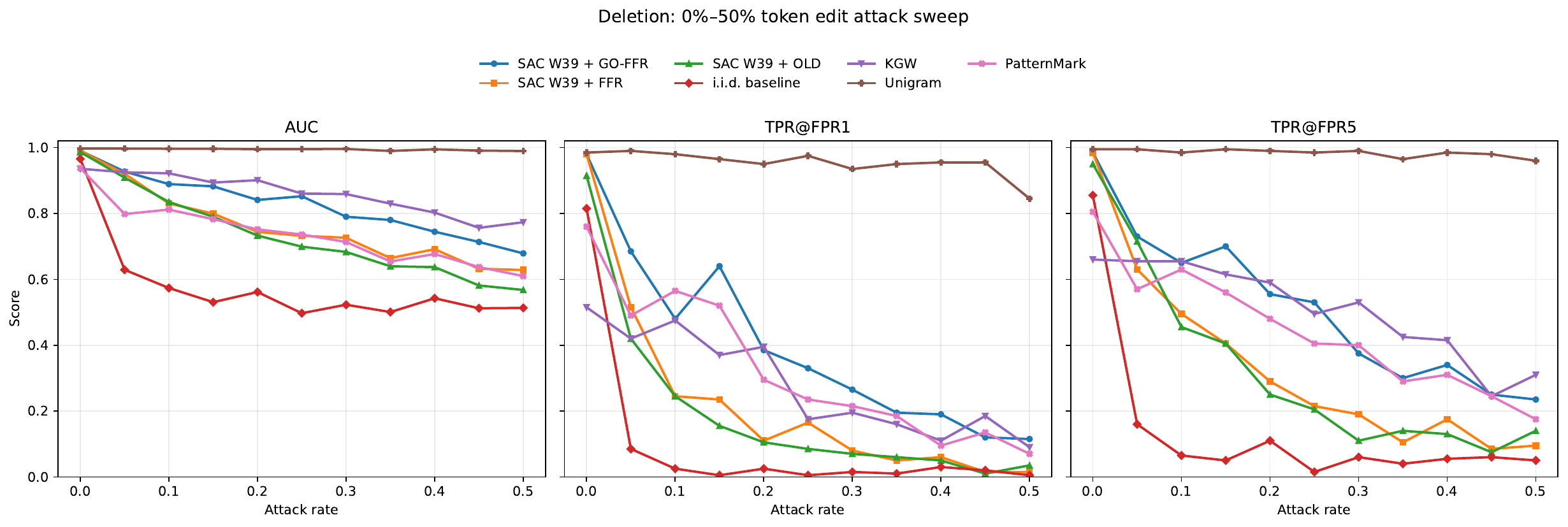}
\caption{Deletion stress-test sweep from 0\% to 50\% token edit rate.  The three panels report AUC, TPR@FPR1, and TPR@FPR5 for SAC-Copula readouts, i.i.d.\ Gumbel, and external clean-text baselines.  The curves are residual-evidence diagnostics: GO-FFR can partially recover coarse offset shifts, while low-FPR TPR remains limited under accumulated deletion drift.}
\label{fig:app-token-edit-sweep-deletion}
\end{figure*}

\begin{figure*}[t]
\centering
\includegraphics[width=0.95\textwidth]{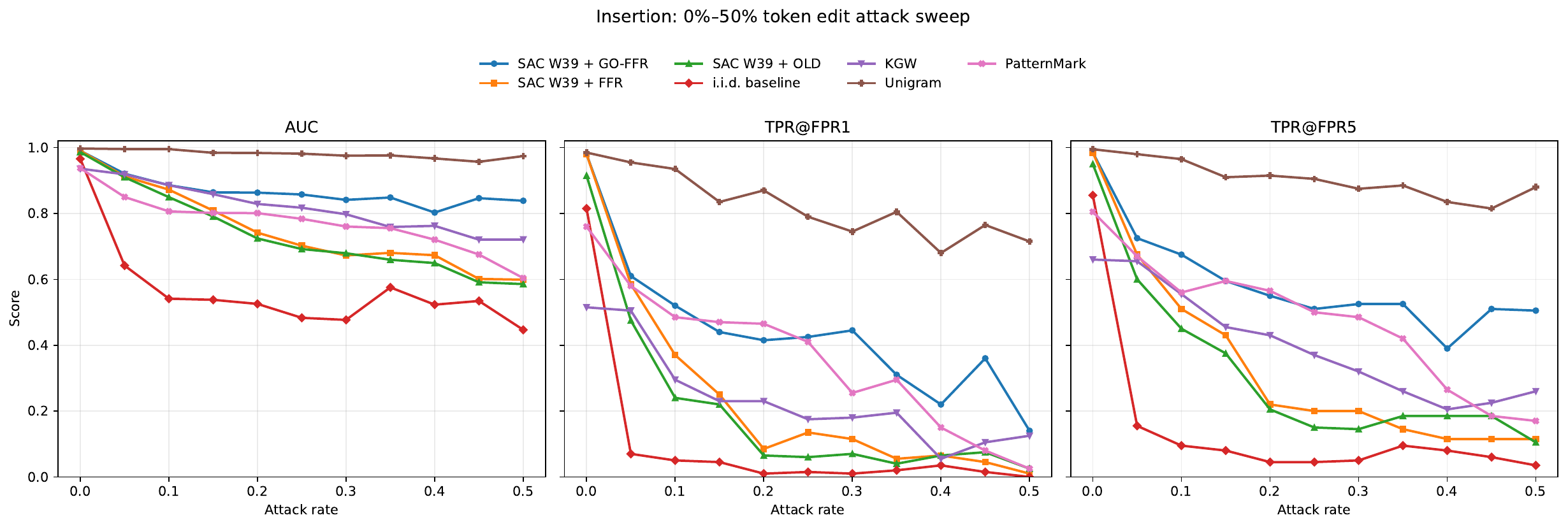}
\caption{Insertion stress-test sweep from 0\% to 50\% token edit rate.  The three panels report AUC, TPR@FPR1, and TPR@FPR5.  Insertion induces synchronization drift by shifting token-to-tape indexing; GO-FFR provides a coarse-offset diagnostic improvement but does not solve cumulative local drift.}
\label{fig:app-token-edit-sweep-insertion}
\end{figure*}

\begin{figure*}[t]
\centering
\includegraphics[width=0.95\textwidth]{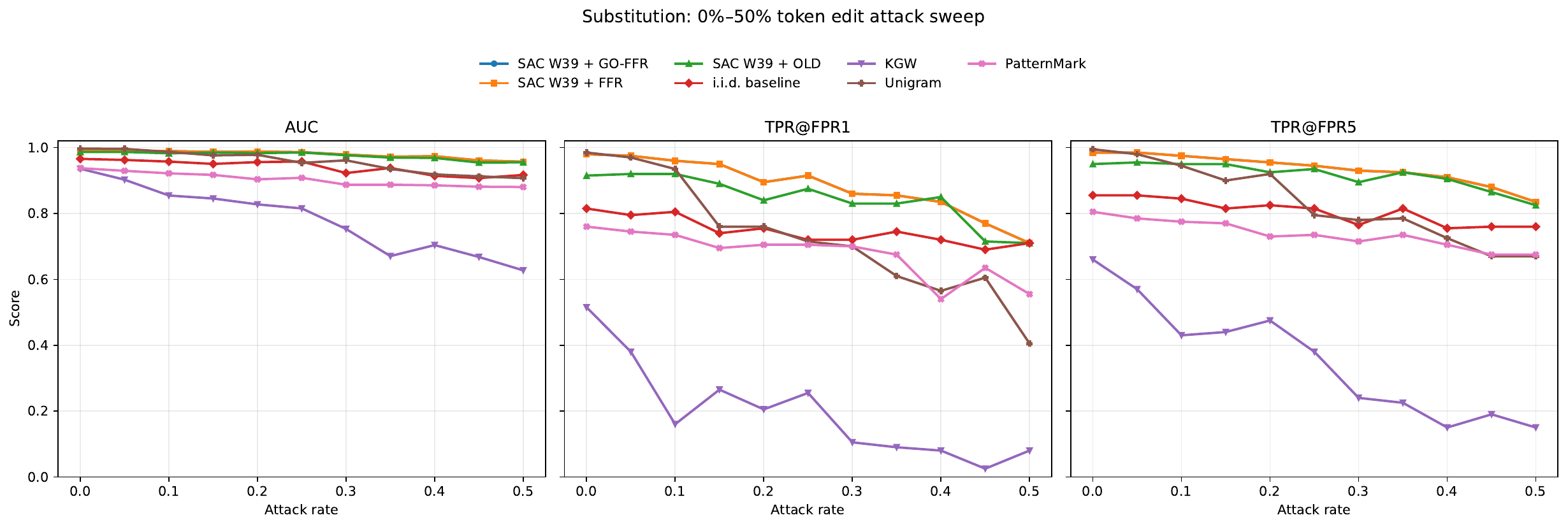}
\caption{Substitution stress-test sweep from 0\% to 50\% token edit rate.  The three panels report AUC, TPR@FPR1, and TPR@FPR5.  Substitution corrupts token identities without shifting all subsequent positions, so it is a controlled token-corruption diagnostic rather than a synchronization-drift setting.}
\label{fig:app-token-edit-sweep-substitution}
\end{figure*}

\clearpage
\endgroup

\subsection{Failure Analysis under Edit Drift}
\label{app:failure-analysis}

\noindent
\begin{minipage}[t]{\columnwidth}
\vspace{0pt}

{\scriptsize
\centering
\setlength{\tabcolsep}{1.2pt}
\begin{tabular}{@{}llrrrr@{}}
\hline
Edit & Detector & AUC & TPR1 & TPR5 & Offset \\
\hline
Deletion & i.i.d. old & 0.513 & 0.005 & 0.050 & 0.000 \\
 & SAC FFR & 0.623 & 0.035 & 0.085 & 0.000 \\
 & SAC GO-FFR & 0.727 & 0.180 & 0.285 & -51.945 \\
\hline
Insertion & i.i.d. old & 0.447 & 0.000 & 0.035 & 0.000 \\
 & SAC FFR & 0.664 & 0.085 & 0.160 & 0.000 \\
 & SAC GO-FFR & 0.878 & 0.335 & 0.550 & 38.135 \\
\hline
\end{tabular}\par
}
\captionof{table}{Diagnostics at 50\% deletion/insertion from the frozen token-edit summary. GO-FFR partially improves coarse synchronization, while cumulative local drift remains.}
\label{tab:app-failure-edit-drift}

Deletion and insertion accumulate synchronization drift. GO-FFR recovers coarse shifts but not cumulative local drift, leaving edit-path alignment unresolved. Evidence comes from the frozen relative-drop, token-attack, and global-offset summaries.

The frozen semantic/paraphrase summaries cover paraphrase results, valid-only filtering, realized TER bins, relative gains, and detector-only comparisons. They show degraded detection under semantic rewriting; offset-style diagnostics provide only partial recovery.

{\scriptsize
\centering
\setlength{\tabcolsep}{1.5pt}
\begin{tabular}{@{}llrrrr@{}}
\hline
Attack & Detector & AUC & TPR1 & TPR5 & Valid \\
\hline
DIPPER R20 & i.i.d. row & 0.580 & 0.075 & 0.105 & 0.545 \\
DIPPER R20 & SAC row & 0.745 & 0.130 & 0.260 & 0.495 \\
DIPPER R30 & i.i.d. row & 0.526 & 0.005 & 0.070 & 0.330 \\
DIPPER R30 & SAC row & 0.633 & 0.020 & 0.140 & 0.320 \\
\hline
\end{tabular}\par
}
\captionof{table}{Semantic/paraphrase failure-analysis rows from the frozen summary, defining the current robustness boundary.}
\label{tab:app-failure-semantic}

\end{minipage}

\subsection{Tail Stability and Correlation-Strength Controls}
\label{app:tail-rho-controls}

The tail analysis in Tables
\ref{tab:app-tail-descriptive}--\ref{tab:app-tail-paired}
re-analyzes 200 frozen LLaDA-8B-Instruct ELI5 outputs for
SAC \(W=39,\sigma_K=15,\rho=0.6\), paired by
\texttt{Prompt\_ID} and \texttt{Gen\_Len}.
Conditional PPL is evaluated with Meta-Llama-3.1-8B-Instruct;
collapse is \(\texttt{Conditional\_PPL}>100\).
Confidence intervals use \(10{,}000\) paired bootstrap resamples
with seed \(20260712\).

\paragraph{Fixed-parameter correlation-strength sweep.}
The canonical 8B sweep varies
\(\rho\in\{0,0.2,0.4,0.6,0.8,1\}\)
while fixing \(W=39\), \(\sigma_K=15\), \(\alpha=1\),
temperature \(1\), generation length \(256\), block length \(32\),
\(128\) diffusion steps, and key \(42\).
Prompts, seeds, decoding, and remasking are fixed; both Old and FFR
detectors are evaluated, and PPL uses Meta-Llama-3.1-8B-Instruct.
The \(\rho=0\) row is the direct matched i.i.d.\ endpoint.
This is a fixed-parameter correlation-strength sweep,
not matched-detectability or watermark-strength optimization.

\begin{table}[!htbp]
\centering
\small
\setlength{\tabcolsep}{3.5pt}
\begin{tabular}{lrrr}
\hline
Metric & Native & i.i.d. & SAC \\
\hline
Mean PPL & 16.840 & 636.069 & 12.163 \\
Median PPL & 6.381 & 8.446 & 8.260 \\
Q95 trimmed mean & 7.201 & 9.864 & 9.333 \\
P90 & 15.387 & 19.603 & 17.911 \\
P95 & 22.580 & 48.420 & 30.810 \\
P99 & 60.653 & 4256.634 & 99.653 \\
Maximum & 1669.349 & 59257.688 & 128.978 \\
PPL\(>100\) & 1/200 & 7/200 & 2/200 \\
\hline
\end{tabular}
\caption{Conditional-PPL summaries for frozen LLaDA--ELI5 outputs.
Q95 is the mean after trimming the highest 5\%.}
\label{tab:app-tail-descriptive}
\end{table}

\begin{table}[!htbp]
\centering
\small
\setlength{\tabcolsep}{2pt}

\begin{tabular}{@{}lr@{}}
\hline
Paired collapse transition & Count \\
\hline
normal \(\rightarrow\) normal & 192/200 \\
i.i.d.\ collapse \(\rightarrow\) SAC normal & 6/200 \\
i.i.d.\ normal \(\rightarrow\) SAC collapse & 1/200 \\
collapse \(\rightarrow\) collapse & 1/200 \\
\hline
\end{tabular}

\vspace{2pt}

\begin{tabular}{@{}lrr@{}}
\hline
Diagnostic & Estimate & 95\% CI \\
\hline
Collapse-rate difference & -0.025 & [-0.050, 0.000] \\
McNemar two-sided & p=0.125 & -- \\
Mean \(\log_{10}\)-PPL difference & -0.0818 & [-0.1664,-0.0063] \\
P99 difference & -4156.981 & [-59177.786,-60.720] \\
Median PPL difference & 0.135 & [-0.741,1.036] \\
SAC lower-PPL win rate & 0.485 & [0.415,0.555] \\
\hline
\end{tabular}

\caption{Paired collapse transitions and statistical diagnostics for SAC relative to matched i.i.d.\ Gumbel. The collapse-rate difference is SAC minus i.i.d.}
\label{tab:app-tail-paired}
\end{table}

\begin{table}[!ht]
\centering
\small
\setlength{\tabcolsep}{3pt}
\begin{tabular}{@{}crrr@{}}
\hline
\(\rho\) &
\shortstack{AUC\\Old / FFR} &
\shortstack{TPR1\\Old / FFR} &
\shortstack{TPR5\\Old / FFR} \\
\hline
0.0 & .9658 / .9683 & .815 / .820 & .855 / .890 \\
0.2 & .9671 / .9753 & .805 / .865 & .895 / .905 \\
0.4 & .9819 / .9861 & .895 / .925 & .950 / .955 \\
0.6 & .9866 / .9900 & .915 / .980 & .950 / .985 \\
0.8 & .9842 / .9845 & .920 / .925 & .950 / .950 \\
1.0 & .9748 / .9631 & .875 / .830 & .940 / .895 \\
\hline
\end{tabular}
\caption{Complete detection results for the canonical six-point \(\rho\) sweep. TPR1 and TPR5 denote TPR at 1\% and 5\% FPR, respectively.}
\label{tab:app-rho-detection}
\end{table}

\begin{table}[!ht]
\centering
\small
\setlength{\tabcolsep}{2.5pt}
\begin{tabular}{@{}crrrrr@{}}
\hline
\(\rho\) &
\shortstack{Median\\PPL} &
P95 &
P99 &
\shortstack{PPL\\\(>100\)} &
SBERT \\
\hline
0.0 & 8.45 & 48.42 & 4256.63 & 3.5\% & .784 \\
0.2 & 8.52 & 36.62 & 2354.28 & 2.5\% & .792 \\
0.4 & 8.28 & 26.00 & 173.26 & 2.0\% & .797 \\
0.6 & 8.26 & 30.81 & 99.65 & 1.0\% & .798 \\
0.8 & 8.17 & 21.97 & 66.18 & 0.5\% & .795 \\
1.0 & 7.38 & 18.00 & 27.55 & 0.0\% & .766 \\
\hline
\end{tabular}
\caption{Complete quality-tail results for the canonical six-point \(\rho\) sweep. At the two reported endpoints for broader quality context, \(\rho=0.6\) has Rep3 \(\approx.414\) and Distinct3 \(\approx.586\), while \(\rho=1\) has Rep3 \(\approx.528\) and Distinct3 \(\approx.472\); unreported intermediate repetition/diversity values are not imputed.}
\label{tab:app-rho-quality}
\end{table}

Detection improves toward moderate correlation and then declines, while the quality dimensions do not move uniformly. The sweep therefore supports \(\rho=0.6\) as a balanced interior operating point, not a universal or global optimum, and does not isolate correlation as a causal theorem.

\FloatBarrier

\subsection{Calibration and Threshold-Transfer Controls}
\label{app:calibration-controls}

\paragraph{Calibration-size sensitivity.}
These controls use LLaDA--ELI5, SAC \(W=39,\sigma_K=15,\rho=0.6\), and model-native \(H_0\).  Calibration sizes are \(50/100/200/500\), with \(20/20/20/1\) repeats, respectively.  Development and evaluation identities are fixed and no evaluation example is used for calibration.

\begingroup
\makeatletter
\setlength{\@dblfptop}{0pt}
\setlength{\@dblfpsep}{10pt}
\setlength{\@dblfpbot}{0pt plus 1fil}
\makeatother
\begin{table}[!htbp]
\centering
\small
\setlength{\tabcolsep}{3pt}
\begin{tabular}{crrr}
\hline
\(N\) & AUC & TPR@1\%FPR & TPR@5\%FPR \\
\hline
50 & \(.9758\pm.0111\) & \(.8025\pm.0715\) & \(.8982\pm.0677\) \\
100 & \(.9867\pm.0034\) & \(.9078\pm.0473\) & \(.9667\pm.0184\) \\
200 & \(.9888\pm.0009\) & \(.9487\pm.0277\) & \(.9812\pm.0060\) \\
500 & .9900 & .9800 & .9850 \\
\hline
\end{tabular}
\caption{ROC discrimination as a function of model-native \(H_0\) calibration size.  Mean and standard deviation are reported when repeated subsamples are available.}
\label{tab:app-cal-size-roc}
\end{table}

\begin{table}[!htbp]
\centering
\small
\setlength{\tabcolsep}{2.5pt}
\begin{tabular}{crrrr}
\hline
\(N\) &
\shortstack{Realized FPR\\@nom1} &
\shortstack{Realized FPR\\@nom5} &
\shortstack{TPR\\@nom1} &
\shortstack{TPR\\@nom5} \\
\hline
50 & 28.88\% & 34.33\% & 98.40\% & 98.65\% \\
100 & 13.03\% & 19.55\% & 98.08\% & 98.55\% \\
200 & 5.13\% & 11.15\% & 97.90\% & 98.53\% \\
500 & 0.50\% & 8.50\% & 98.00\% & 98.50\% \\
\hline
\end{tabular}
\caption{Operational-threshold diagnostic across calibration sizes.  Strong ROC discrimination does not imply insensitivity of fixed low-FPR thresholds to \(H_0\) sample size.}
\label{tab:app-cal-size-threshold}
\end{table}

\paragraph{Calibration-source controls.}
All four source conditions use \(N=500\), the same development identities, the same 200 LLaDA--ELI5 Native negative evaluations, and the same 200 LLaDA--ELI5 SAC positive evaluations; only the \(H_0\) source changes.

\begin{table}[!htbp]
\centering
\small
\setlength{\tabcolsep}{3pt}
\begin{tabular}{lrrr}
\hline
\(H_0\) source & AUC & TPR@1\%FPR & TPR@5\%FPR \\
\hline
LLaDA--ELI5 Native & .9900 & .980 & .985 \\
ELI5 human answer & .9878 & .955 & .980 \\
C4 raw text & .9871 & .955 & .975 \\
Wikipedia raw text & .9861 & .950 & .970 \\
\hline
\end{tabular}
\caption{Four-source \(H_0\) calibration control with fixed development and evaluation identities.  The results quantify source sensitivity; no formal equivalence margin was pre-specified.}
\label{tab:app-cal-source}
\end{table}

\begin{table}[!htbp]
\centering
\small
\setlength{\tabcolsep}{2.5pt}
\begin{tabular}{llrrr}
\hline
\shortstack{Calibration\\\(H_0\)} &
\shortstack{Negative\\Eval} &
AUC & TPR1 & TPR5 \\
\hline
Native & Native & .9900 & .9800 & .9850 \\
Native & ELI5 human & .9908 & .9850 & .9850 \\
ELI5 human & Native & .9878 & .9550 & .9800 \\
ELI5 human & ELI5 human & .9902 & .9850 & .9900 \\
\hline
\end{tabular}
\caption{Native/Human \(2\times2\) ROC control.  This matrix reports score discrimination rather than realized FPR at frozen operational thresholds.}
\label{tab:app-cal-native-human}
\end{table}

\paragraph{Threshold portability and prompt shift.}
Using Wikipedia \(H_0\) thresholds on ELI5 Native evaluation gives realized FPR \(6.5\%\) (\(13/200\)) and TPR \(97.5\%\) at the nominal 1\% point, and realized FPR \(21.0\%\) (\(42/200\)) and TPR \(98.5\%\) at the nominal 5\% point.  Thus, strong ROC separation is not equivalent to threshold portability.

\par\smallskip
\noindent
\begin{minipage}{\columnwidth}
\centering
{\scriptsize
\setlength{\tabcolsep}{1.5pt}
\begin{tabular}{@{}c p{0.45\columnwidth} rrr@{}}
\hline
Condition & \(H_0\) construction & AUC & TPR1 & TPR5 \\
\hline
A & Original 500 ELI5 questions + original starting prompt & .9900 & .980 & .985 \\
B & Disjoint 500 ELI5 questions + same starting prompt & .9889 & .970 & .985 \\
C & Same new questions/seeds + alternative starting prompt & .9907 & .980 & .985 \\
\hline
\end{tabular}
}
\captionof{table}{Same-model prompt-shift control.  A\(\rightarrow\)B changes question/content samples; B\(\rightarrow\)C changes the prompt template with questions and seeds fixed; A\(\rightarrow\)C is the combined tested same-model/same-task shift.  This is not a prompt-invariance claim beyond the evaluated setting.}
\label{tab:app-cal-prompt-shift}
\end{minipage}
\par\smallskip

\paragraph{Frozen-threshold C4 \texttt{realnewslike} FPR.}
This detector-only transfer test freezes the original LLaDA--ELI5 model-native-\(H_0\) thresholds, performs no C4 refit, and evaluates \(2{,}000\) held-out non-watermarked C4 \texttt{realnewslike} records.  It is not C4 watermark generation and is distinct from the C4-en continuation experiments below.

\begin{table}[!htbp]
\centering
\small
\setlength{\tabcolsep}{3pt}
\begin{tabular}{rrrr}
\hline
Nominal FPR & Realized FPR & Exact 95\% CI & SAC TPR \\
\hline
1\% & 0.95\% & [0.57\%, 1.48\%] & .980 \\
5\% & 5.50\% & [4.54\%, 6.59\%] & .985 \\
\hline
\end{tabular}
\caption{Frozen-threshold false-positive transfer from LLaDA--ELI5 Native \(H_0\) to held-out C4 \texttt{realnewslike} negatives.}
\label{tab:app-frozen-c4-fpr}
\end{table}

As a secondary diagnostic, rebuilding \(H_0\) on C4 \texttt{realnewslike} yields realized FPR \(2.95\%\), exact 95\% CI \([2.25\%,3.79\%]\), at nominal 1\%, and \(7.55\%\), CI \([6.43\%,8.80\%]\), at nominal 5\%.  This evaluated sample does not show improved false-positive control after rebuilding \(H_0\) on C4; it does not establish that target-domain recalibration is generally harmful.  Frozen-threshold realized FPR and ROC-selected detection answer different operational questions.

\FloatBarrier
\endgroup
\subsection{Targeted Backbone and Task Transfer}
\label{app:targeted-transfer}

\paragraph{Dream-v0-Instruct-7B on ELI5.}
This targeted second-backbone evaluation uses 500 held-out ELI5 human answers for calibration, a separate 200 human answers for negative evaluation, and 200 method-specific Dream generations for positive evaluation.  The i.i.d.\ system uses its original exact-tape detector, while SAC uses FFR.

\begin{table}[!htbp]
\centering
\small
\setlength{\tabcolsep}{2.5pt}
\begin{tabular}{lrrrr}
\hline
\multicolumn{5}{c}{Detection} \\
\hline
Method & AUC & TPR1 & TPR5 & \\
\hline
i.i.d.\ Gumbel & 1.000000 & 1.000 & 1.000 & \\
SAC-Copula & .998625 & .995 & .995 & \\
\hline
\multicolumn{5}{c}{} \\
\hline
\multicolumn{5}{c}{Quality} \\
\hline
Source & Median & P95 & P99 & \shortstack{Composite\\collapse} \\
\hline
ELI5 human answers & 13.42 & 24.04 & 30.55 & 0.0\% \\
i.i.d.\ Gumbel & 30.02 & 889.58 & 5988.25 & 24.5\% \\
SAC-Copula & 17.84 & 81.69 & 240.68 & 4.0\% \\
\hline
\end{tabular}
\caption{Dream-7B detection and quality evidence on ELI5.  Composite collapse is the Dream experiment's preregistered PPL/repetition rule and is not the LLaDA \(\texttt{Conditional\_PPL}>100\) statistic.}
\label{tab:app-dream-transfer}
\end{table}

Both methods retain near-perfect detection in this protocol, while SAC reduces the i.i.d.\ upper-tail and composite-collapse profile.  This provides targeted evidence of cross-DLM generalization to a second backbone.

\paragraph{LLaDA on C4-en continuation.}
The task uses a continuation instruction with a 128-token passage and generates 256 tokens.  SAC uses \(W=39,\sigma_K=15,\rho=0.6\) with no C4-specific watermark retuning.  Calibration uses 200 C4-prompted unwatermarked LLaDA outputs; evaluation uses 500 raw non-watermarked C4-en records and 200 method-specific watermarked outputs.  The method-matched pipelines use Old for i.i.d.\ and FFR for SAC.
\begin{table}[!htbp]
\centering
\small
\setlength{\tabcolsep}{3pt}

\begin{tabular}{@{}lrrr@{}}
\hline
\multicolumn{4}{c}{Detection} \\
\hline
Pipeline & AUC & TPR1 & TPR5 \\
\hline
i.i.d.+Old & .8260 & .415 & .530 \\
SAC+FFR & .9138 & .670 & .805 \\
\hline
\end{tabular}

\vspace{2pt}

\begin{tabular}{@{}lrrr@{}}
\hline
\multicolumn{4}{c}{Quality} \\
\hline
Method & Median PPL & P99 & PPL\(>100\) \\
\hline
i.i.d. & 13.71 & 35740.32 & 10.5\% \\
SAC & 13.73 & 494.57 & 3.0\% \\
\hline
\end{tabular}

\caption{Method-matched LLaDA C4-en continuation results.
Detection differences are attributed to the complete i.i.d.+Old and
SAC+FFR pipelines, not solely to generation-side SAC.}
\label{tab:app-c4-method-matched}
\end{table}

\FloatBarrier

The two pipelines have nearly identical median PPL, while SAC+FFR has a substantially less severe quality tail and stronger detection under the submitted method-matched protocol.  This provides targeted evidence of cross-domain generalization to a second task/source setting.

\paragraph{Separate common-detector control.}
This is a distinct detector-control protocol and is not merged with the method-matched result above.

\begin{table}[!htbp]
\centering
\small
\setlength{\tabcolsep}{2pt}

\begin{tabular}{@{}lrrrr@{}}
\hline
\multicolumn{5}{c}{Detection control} \\
\hline
Method &
\shortstack{Old\\AUC} &
\shortstack{FFR\\AUC} &
\shortstack{FFR\\TPR1} &
\shortstack{FFR\\TPR5} \\
\hline
i.i.d. & .9107 & .8627 & .130 & .315 \\
SAC & .9236 & .9221 & .785 & .825 \\
\hline
\end{tabular}

\vspace{2pt}

\begin{tabular}{@{}lrrrrr@{}}
\hline
\multicolumn{6}{c}{Quality} \\
\hline
Method & Median & P95 & P99 &
\shortstack{PPL\\\(>100\)} & SBERT \\
\hline
i.i.d. & 13.71 & 680.48 & 35740.32 & 10.5\% & .570 \\
SAC & 13.73 & 78.61 & 494.57 & 3.0\% & .601 \\
\hline
\end{tabular}

\caption{Separate C4 common-detector audit.  These detector-control values must not be interpreted as the submitted method-matched C4 result in Table~\ref{tab:app-c4-method-matched}.}
\label{tab:app-c4-common-detector}
\end{table}

Additional quality context in this separate audit is Rep3 \(0.7298\rightarrow0.6954\), Distinct3 \(0.2702\rightarrow0.3046\), and Ent3 \(1.7293\rightarrow1.9986\) from i.i.d.\ to SAC.  The C4 raw-text calibration-source control, frozen-threshold C4 \texttt{realnewslike} FPR test, C4-en method-matched experiment, and this common-detector audit retain their distinct \(H_0\), evaluation, denominator, and detector identities.

\FloatBarrier

\end{document}